%% file: main.tex
\documentclass[10pt,twocolumn,letterpaper]{article}

\usepackage[pagenumbers]{cvpr} % To force page numbers, e.g. for an arXiv version


\definecolor{cvprblue}{rgb}{0.21,0.49,0.74}
\usepackage[pagebackref,breaklinks,colorlinks,citecolor=cvprblue]{hyperref}

\usepackage[ruled,vlined]{algorithm2e}
\usepackage{algorithmic}
\usepackage{float}
\usepackage[dvipsnames]{xcolor}

\newcommand{\redt}[1]{\textcolor{red}{\textbf{#1}}}

\newcommand{\greent}[1]{\textcolor{ForestGreen}{\textbf{#1}}}

\SetKwInput{KwInput}{Input}
\SetKwInput{KwOutput}{Output}

\newcommand{\inst}[1]{\textsuperscript{#1}}

\usepackage{wrapfig,lipsum,booktabs}

\def\paperID{****} % *** Enter the Paper ID here
\def\confName{CVPR}
\def\confYear{2025}

\title{Meta-Learned Modality-Weighted Knowledge Distillation for Robust Multi-Modal Learning with Missing Data}

\author{\textbf{Hu Wang\inst{a}, Salma Hassan\inst{a}, Yuyuan Liu\inst{g}, Congbo Ma\inst{c}, Yuanhong Chen\inst{b}}, \\
\textbf{Qing Li\inst{a}, Jiahui Geng\inst{a}, Bingjie Wang\inst{e}, Yu Tian\inst{f}, Yutong Xie\inst{a}, Jodie Avery\inst{b}}, \\
\textbf{Louise Hull\inst{b}, Ian Reid\inst{a}, Mohammad Yaqub\inst{a}, Gustavo Carneiro\inst{d}} \\
\inst{a}Mohamed bin Zayed University of Artificial Intelligenc, UAE \\
\inst{b}University of Adelaide, Australia 
\inst{c}New York University Abu Dhabi, UAE \\
\inst{d}University of Surrey, UK 
\inst{e}Boai hospital of Zhongshan, China \\
\inst{f}University of Central Florida, USA 
\inst{g}University of Oxford, UK \\
}

\begin{document}
\maketitle

\begin{abstract}
In multi-modal learning, some modalities are more influential than others, and their absence can have a significant impact on classification/segmentation accuracy. Addressing this challenge, we propose a novel approach called \textbf{Meta}-learned Modality-weighted \textbf{K}nowledge \textbf{D}istillation (MetaKD), which enables multi-modal models to maintain high accuracy even when key modalities are missing. MetaKD adaptively estimates the importance weight of each modality through a meta-learning process. These learned importance weights guide a pairwise modality-weighted knowledge distillation process, allowing high-importance modalities to transfer knowledge to lower-importance ones, resulting in robust performance despite missing inputs. Unlike previous methods in the field, which are often task-specific and require significant modifications, our approach is designed to work in multiple tasks ($e.g.$, segmentation and classification) with minimal adaptation. Experimental results on five prevalent datasets, including three Brain Tumor Segmentation datasets (BraTS2018, BraTS2019 and BraTS2020), the Alzheimer's Disease Neuroimaging Initiative (ADNI) classification dataset and the Audiovision-MNIST classification dataset, demonstrate the proposed model is able to outperform the compared models by a large margin. The code is available at \url{https://github.com/billhhh/MetaKD}.
\end{abstract}

\section{Introduction}

Multi-modal learning builds classification and segmentation models that integrate and process information from multiple data modalities ($e.g.$, text, images, audio, video, sensor data, etc.). 
The goal of multi-modal learning is to leverage complementary information from different modalities to enhance performance when compared to using each modality in isolation.
This approach has been explored in many fields, such as computer vision (CV) and medical image analysis (MIA), and applications, like autonomous driving~\cite{prakash2021multi}, 
robot navigation~\cite{wang2020soft}, general multi-modal dialog system~\cite{openai2023gpt4}, and 
medical multi-modal diagnosis~\cite{wang2022uncertainty}.
However, a significant challenge in multi-modal learning is dealing with scenarios where some input modalities are missing, as many existing models assume that all modalities are always available, limiting their robustness in real-world applications.

\begin{figure*}[t]
\centering
\includegraphics[width=0.8\textwidth]{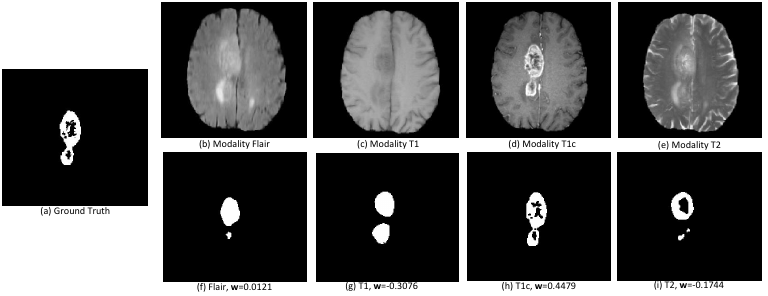}
\caption{Visualization of the enhancing tumor (ET) segmentation results from each modality on a BraTS2018 image. Sub-figure (a) shows the Ground Truth segmentation; (b)-(e) and (f)-(i) display the original modalities (Flair, T1, T1c or T2) and their segmentations, along with the importance weights ($\mathbf{w}$ in the legend) learned automatically (before normalization) by our MetaKD. Note that as the importance weight increases, segmentation accuracy improves. This insight enables the development of our meta-learning approach that automatically distills knowledge from higher-accuracy modalities to those with lower accuracy, enhancing overall model performance.
}
\label{fig:vis-iwv}
\end{figure*}

The challenge presented by multi-modal learning with missing modalities has attracted significant interest.
In CV, approaches like learning a shared multi-modal space~\cite{yin2017unified} and SMIL~\cite{ma2021smil} aim to address this issue. In MIA, some models~\cite{havaei2016hemis,dorent2019hetero} use statistical information (mean and variance) for decoding missing modalities, while others~\cite{shen2019brain,wang2021acn} estimate missing features from available ones. Wang et al.~\cite{wang2021acn} introduced a ``dedicated'' strategy, training independent models for each missing modality scenario. Another approach focuses on feature disentanglement, separating modality-shared and modality-specific features~\cite{chen2019robust,wang2023multi}. 
A fact in multi-modal learning, which is scarcely explored by previous methods, is that there are certain modalities that contribute more than others for certain tasks, as shown in Fig.~\ref{fig:vis-iwv}. 
Moreover, many existing approaches are task-specific, making them difficult to adapt across tasks like classification and segmentation.

In this paper, we propose a novel method called \textbf{Meta}-learned Modality-weighted \textbf{K}nowledge \textbf{D}istillation (MetaKD) for multi-modal learning with missing modalities designed to handle the performance and task adaptation challenges mentioned above. MetaKD leverages meta-learning to automatically estimate modality importance weights and distill knowledge from the most ``important'' modalities. An example is shown in Fig.~\ref{fig:vis-iwv}, where higher segmentation accuracy is automatically associated with a larger importance weight by MetaKD. The learned importance is adopted to weight the knowledge distillation terms to distill information from modalities with higher importance to those with lower importance. 
In addition, MetaKD is adaptable to multiple tasks, such as classification and segmentation. Our main contributions are:
\begin{itemize}
\item An innovative multi-modal learning model named MetaKD, which handles missing modalities by distilling knowledge from higher-accuracy modalities to lower-accuracy ones using meta-learning; and 
\item A flexible model design to
enable MetaKD to easily adapt to multiple tasks, such as classification and segmentation.
\end{itemize}
Experimental results on 5 prevalent benchmarks demonstrate that MetaKD achieves state-of-the-art performance.

\begin{figure*}[t]
\centering
\includegraphics[width=0.8\textwidth]{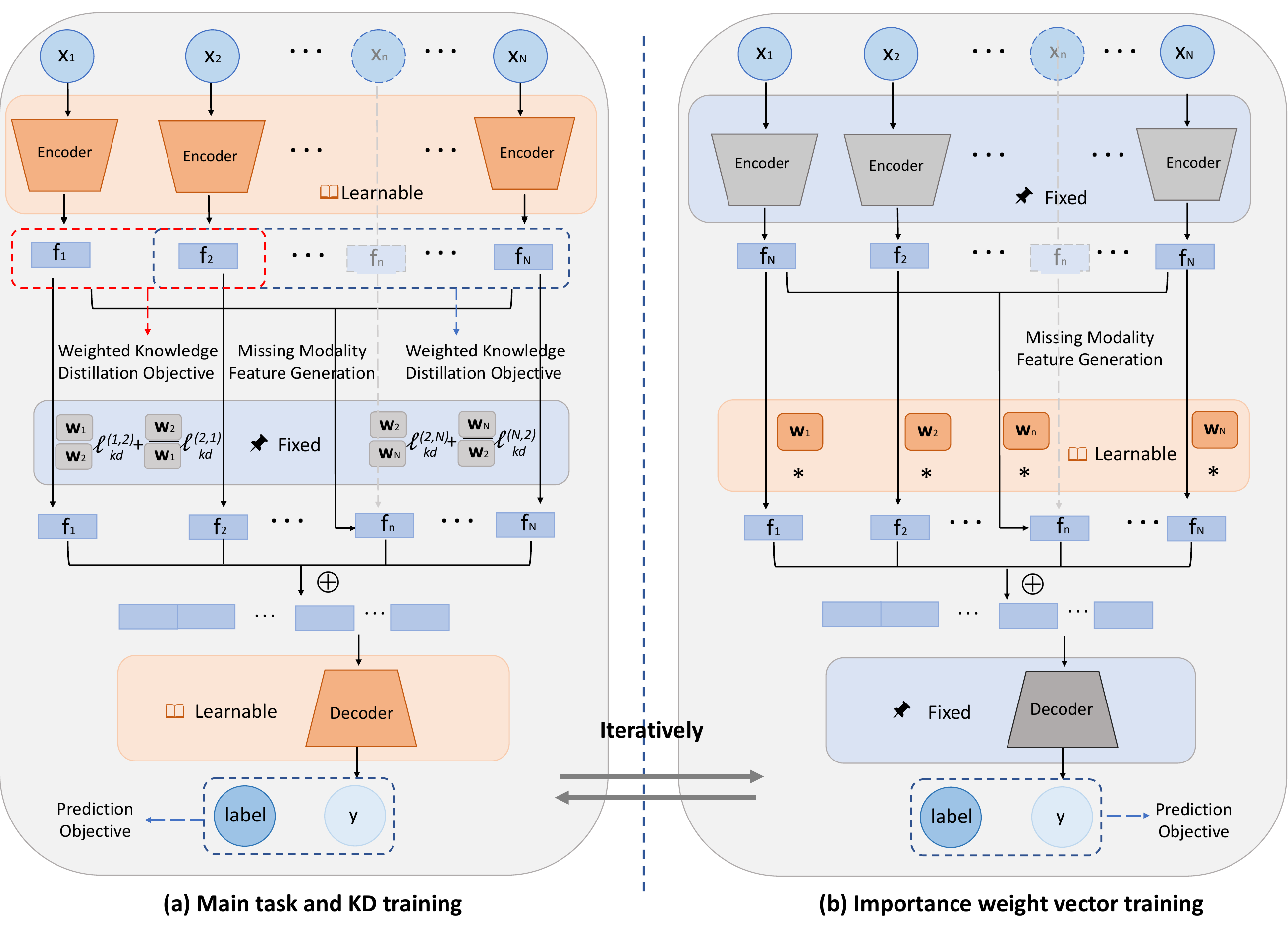}
\caption{MetaKD model framework for (a) main task and modality-weighted knowledge distillation training; (b) importance weight vector training. The two processes will be performed iteratively.
}
\label{fig:framework}
\end{figure*}

\section{Related Work}
\subsection{Multi-modal Learning}
\label{sec:multi_modal_learning_lit_review}

Multi-modal learning is essential in various applications, such as medical image analysis, computer vision, natural language processing, and robotics, where data from different sources provides a complementary understanding for improved task performance. These algorithms extract features from each modality and combine them to enhance decision-making or predictions.

In MIA, Dou et al.~\cite{dou2020unpaired} introduced a chilopod-shaped architecture optimized using a modality-dependent feature normalization and a knowledge distillation objective, which only works effectively when all modalities are available in the input.  Monteiro et al.~\cite{monteiro2020stochastic} developed a method based on pixel-wise coherency, modeling aleatoric uncertainty in image segmentation with the generation of multiple hypotheses through joint distributions over label maps. Han et al.~\cite{han2021trusted} designed a trusted multi-view classifier by modeling multi-modal uncertainties using the Dirichlet distribution and fusing features through Dempster's rule. Wang et al.~\cite{wang2022uncertainty} introduced an uncertainty-aware multi-modal learning model based on cross-modal random network prediction.

In CV, channel exchanging and multi-modal learning were combined to fuse features effectively~\cite{wang2020deep}.  For video/audio classification and retrieval tasks, self-supervised learning was proposed~\cite{patrick2020multi,patrick2021space} to train multi-modal models on additional data, leading to large performance improvements. A model to enhance video-and-sound source localization accuracy was developed~\cite{chen2021localizing} by defining a trusted tri-map middle-ground. Jia et al.~\cite{jia2020semi} introduced a model for multi-view learning by constraining the view-specific features to be orthogonal to the view-shared features. Feature disentanglement methods~\cite{lee2018diverse,liu2022learning} aim to model data variations by learning modular and informative representations. More recently, VL-Mamba~\cite{qiao2024vl} explores the use of state space models for multimodal learning by introducing a vision-language connector with a Vision Selective Scan module, enabling efficient integration of visual features into the Mamba language model. Although these approaches achieve effective multi-modal learning, they assume all modalities are available during training and testing, limiting their practical use in scenarios where some modalities may be missing.

\subsection{Multi-modal Learning with Missing Modalities}

To address the issue of missing modalities in multi-modal learning, various methods have been developed~\cite{ma2022multimodal,lee2023multimodal,lin2023missmodal,liu2023attention}. In MIA, the HeMIS model was developed~\cite{havaei2016hemis} to learn modality-specific embeddings, allowing statistical information (e.g., mean and variance) on these embeddings to produce segmentations.  Dorent et al.~\cite{dorent2019hetero} extended HeMIS with a multi-modal variational auto-encoder (MVAE) that produced pixel-wise classifications from these embeddings. Similarly, auto-encoder structures have been used for unsupervised learning of missing modalities~\cite{chartsias2017multimodal,van2018learning}. Several other approaches~\cite{shen2019brain,wang2021acn,hu2020knowledge} proposed the learning of missing modality features from full modality models to enhance embeddings, while feature disentanglement methods~\cite{chen2019robust,wang2023multi} separated shared and modality-specific features for better performance. A vision transformer model~\cite{zhang2022mmformer} was also introduced for multi-modal brain tumor segmentation, aiming to fuse features from all modalities into comprehensive representations.

In CV, Yin et al.~\cite{yin2017unified} proposed a model to learn a unified subspace for incomplete multi-view data, while Albanie et al.~\cite{albanie2018emotion} developed a cross-modal transfer model to transfer knowledge between modalities. However, it cannot learn the modality importance automatically and only works for video-speech classification. Liu et al.~\cite{liu2024modality} proposes a modality translation framework to address multimodal sentiment analysis under uncertain missing modalities, enabling effective representation recovery and robust prediction even when some modalities are unavailable. Other models~\cite{yang2023learning,zhou2021feature,azad2022smu,chen2023disentangle,konwer2023enhancing} addressed missing modalities, but were often task-specific or restricted to particular diseases, limiting broader applicability. Ma et al.~\cite{ma2021smil} proposed the SMIL model, which employed a meta-learning algorithm to reconstruct features from missing modalities. Unlike SMIL, which uses meta-learning to handle missing modalities by updating two auxiliary networks, we apply meta-learning to train a set of modality importance weights for modality-weighted knowledge distillation\footnote{Also, we notice that in practice, SMIL has high memory complexity and exhibits unstable performance in missing modality scenarios.}.

Similarly to the multi-modal methods presented Sec.~\ref{sec:multi_modal_learning_lit_review}, the solutions mentioned above are task-specific and often neglect the fact that some modalities are more informative than others. 
Our proposed MetaKD aims to address these gaps by automatically identifying and distilling the knowledge from influential modalities to dramatically enhance the model performance, particularly when such influential modalities are not present in the input data. MetaKD is also designed to easily adapt to disparate tasks, such as classification or segmentation.

\section{Methodology}
\subsection{Model Structure}

Let us denote the $N$-modality data as $\mathcal{M}_l = \{\mathbf{x}_i^l\}_{i=1}^{N}$, where $\mathbf{x}_i^l \in \mathcal{X}$ represents the $l^{th}$ data sample and the subscript $i$ indexes the modality. The corresponding label for each set $\mathcal{M}$ is represented by $\mathbf{y}^{l} \in \mathcal{Y}$, where $\mathcal{Y}$ denotes the ground-truth annotation space. 
To simplify the notation, we may omit the subscript $l$ when the context is clear. 

Our proposed MetaKD, illustrated in Fig.~\ref{fig:framework}, consists of a two-stage meta-learning method.
The first stage, shown in Fig.~\ref{fig:framework}(a), consists of multiple teacher-student training processes, along with the main task optimization (e.g. classification or segmentation), where for each pair of available modalities, we distill the knowledge from the teacher 
to the student using the ratio of their importance weights. The second stage, shown in Fig.~\ref{fig:framework}(b), estimates the importance weight for each modality when training for the task represented by the label $\mathbf{y}$ (note that it is possible that a single dataset has multiple sub-tasks, 
such as the three types of tumor segmentation in the BraTS2018 dataset~\cite{menze2014multimodal,bakas2018identifying}).
This modality importance weight indicates the ``amount'' of knowledge contained in that modality.

As illustrated in Fig. \ref{fig:framework}, each non-missing modality $\mathbf{x}_i$ is fed into the encoder, parameterized by $\theta_i$, which extracts the feature $\mathbf{f}_i$, with
\begin{equation} \small
\label{eq:enc}
        \mathbf{f}_i = f_{\theta_i}(\mathbf{x}_i).
\end{equation}
In order to simplify the model, the encoders' parameters $\{\theta_i\}_{i=1}^{N}$ are shared among all modalities. 
For the missing modalities in Eq.~\eqref{eq:enc}, let us assume that the modalities in $\mathcal{Q} \subset \mathcal{M}$ are unavailable (i.e., all $\mathbf{x}_n \in \mathcal{Q}$ are represented by $\mathbf{x}_n = \emptyset$), then the missing features from those modalities can be generated based on the available features, as follows: 
\begin{equation} \small
    \mathbf{f}_n = \frac{1}{N-|\mathcal{Q}|}\sum_{\mathbf{x}_i \in \mathcal{M} \setminus \mathcal{Q}} f_{\theta_i}(\mathbf{x}_i)
    \label{eq:ftgen}
\end{equation}
for all $\mathbf{x}_n \in \mathcal{Q}$, with $\setminus$ denoting the set subtraction operator.

The features obtained from Eq.~\eqref{eq:enc} or generated from Eq.~\eqref{eq:ftgen} are concatenated and fed into the decoder,  parameterized by $\zeta$, to make predictions, as in:
\begin{equation} \small
\label{eq:dec}
        \hat{\mathbf{y}} = f_{\zeta}\left(\oplus_{i=1}^N \mathbf{f}_i\right),
\end{equation}
where $\hat{\mathbf{y}}$ is the prediction, and 
$\oplus_{i=1}^N$ is the concatenator operator of the $N$ extracted features.

\subsection{Knowledge Distillation Meta-learning}

We formulate the training of MetaKD with a meta-learning bi-level optimization that contains a meta-parameter represented by an \textbf{importance weight vector (IWV)} $\mathbf{w}=\left[\mathbf{w}_1, \ldots, \mathbf{w}_N\right]^{\top} \in \mathbb{R}^N$. 
The inner learning uses the training set $\mathcal{D}_t = \{ (\mathcal{M},\mathbf{y})^l \}_{l=1}^{|\mathcal{D}_t|}$ to optimize the parameters of the model responsible to realize the main task of interest (e.g., classification or segmentation), while the meta-parameters are estimated from the outer learning that optimizes a meta-objective with a validation set $\mathcal{D}_v = \{ (\mathcal{M},\mathbf{y})^l \}_{l=1}^{|\mathcal{D}_v|}$ (where $\mathcal{D}_t \bigcap \mathcal{D}_v = \emptyset$, i.e., $\mathcal{D}_t$ and $\mathcal{D}_v$ are extracted from the same original training set, but they are mutually exclusive), as in:
\begin{equation} \small
\label{eq:meta}
\scalebox{1.0}{$
\begin{aligned}
\mathbf{w}^* & =  \arg \min _{\mathbf{w}} \sum_{(\mathcal{M}, \mathbf{y}) \in \mathcal{D}_v} \ell_{meta} \left(\mathbf{y}, f_{\zeta^{*}}\left(\oplus_{i = 1}^{N} \mathbf{w}_i \times \mathbf{f}_i^* \right)\right) \\
\text { s.t. } &  \theta^*,\zeta^* =  \arg \min _{\theta,\zeta} \sum_{(\mathcal{M}, \mathbf{y}) \in \mathcal{D}_t} \Bigg [ \ell_{task} \left(\mathbf{y}, 
f_{\zeta}\left(\oplus_{i=1}^N \mathbf{f}_{i} \right)\right) \\
& \;\;\;\;\;\;\;\;\;\;\;\;\;\;\;\;\;\;\;\;\;\;\;\;\;\;\;\;\;\;\;\;+ \alpha  \sum_{i,j = 1}^{N}   \frac{\mathbf{w}_i^*}{\mathbf{w}_j^*} \times 
\ell_{kd}^{(i,j)}\left(  \mathbf{f}_{i}, \mathbf{f}_{j} \right) 
\Bigg ],
\end{aligned}
$}
\end{equation}
where $\mathbf{f}^*_i = f_{\theta_i^*}\left(\mathbf{x}_i\right)$ or computed with~\eqref{eq:ftgen} (similarly for $\mathbf{f}_i = f_{\theta_i}\left(\mathbf{x}_i\right)$),
$\theta$ and $\zeta$ are the learnable parameters of the model encoder and decoder, respectively, 
the loss functions $\ell_{meta}(.)$ and $\ell_{task}(.)$ are represented by the same element-wise (pixel in 2D or voxel in 3D) cross-entropy loss plus Dice loss for segmentation and the cross-entropy loss for classification.
Also in Eq.~\eqref{eq:meta}, we have the \textbf{modality-weighted knowledge distillation} loss defined as
\begin{equation} \small
\label{eq:kd}
    \ell_{kd}(\mathbf{f}_i,\mathbf{f}_j) = \delta(\mathbf{x}_i \ne \emptyset) \times \delta(\mathbf{x}_j \ne \emptyset)  \times \| \mathbf{f}_i - \mathbf{f}_j \|_p,
\end{equation}
which is the $p$-norm of the difference between the encoder features of modalities $i,j \in \{1,...,N\}$ (with $\delta(\mathbf{x}_i \ne \emptyset)$ indicating that the modality $i$ is present), with 
$\alpha$ being the trade-off factor between the $\ell_{task}(.)$ and $\ell_{kd}(.)$ to balance the contribution of both losses to the modality-weighted knowledge distillation objective, and $\frac{\mathbf{w}_i^*}{\mathbf{w}_j^*}$ denoting the cost to penalize feature differences particularly with respect to the features from the modalities with large IWV. We need to normalize the importance weights for numerical stability; if not normalized, the scale differences between importance elements may be learned to be huge as there is no constraint on it. If we use ReLU or sigmoid normalization for each element within IWV, the elements lose connection with each other as ReLU or sigmoid do not take all elements into consideration, which will, in turn, affect the gradient back-propagation. The influences of different normalization methods on model performance can be found in the Supplementary Material. 
Thus, we normalize the IWV weight vector with a softmax activation function which provides the best performance by keeping the relative importance of the modalities, with
\begin{equation} \small
\label{eq:IWV_norm}
    \mathbf{w}_i \leftarrow 
    e^{\mathbf{w}_i} \div \left ( \sum_{j=1}^N e^{\mathbf{w}_j} \right ), \ \text { for } i=1, \ldots, N,
\end{equation}
updated at each iteration of the optimization in Eq.~\eqref{eq:meta}.
This knowledge distillation between pairs of teachers and students leads to a convergence of features to be close to the features in the feature space from the modalities with greater weight in the IWV.
Additionally, note that 
such pairwise knowledge distillation allows 
each modality to act as a teacher to other modalities (with different importance weights), resulting in the distillation of knowledge from potentially many modalities through this dynamical process.
Details of the training algorithm for MetaKD are shown in the supplementary material.

\section{Experiments}
Experimental settings, implementation details of 5 datasets, as well as extra results for the BraTS2019/BraTS2020 datasets and analytical experiments can be found in the supplementary material.

\subsection{Model Performance on Medical Image Segmentation}

\begin{table*}[h]\small
\setlength\tabcolsep{1pt}
\begin{center}
\caption{Model performance comparison on the test set of \textbf{segmentation} Dice score (in \%) on BraTS2018 of \textbf{non-dedicated training}. Our models are compared with U-HeMIS (HMIS)~\cite{havaei2016hemis}, U-HVED (HVED)~\cite{dorent2019hetero}, Robust-MSeg (RbSeg)~\cite{chen2019robust}, mmFormer (mmFm)~\cite{zhang2022mmformer} and ShaSpec (ShSpc)~\cite{wang2023multi}. ``Imprv'' denotes the improvement (in percentage) between our proposed MetaKD model and the best of all other models. The best and second best results for each column within a certain type of tumor are in \textbf{bold} and \textit{Italic}, respectively.}\label{tab:brats2018}
\resizebox{1.0\linewidth}{!}{
\begin{tabular}{|cccc|ccccc|cc|ccccc|cc|ccccc|cc|}
\hline
\multicolumn{4}{|c|}{Modalities}               & \multicolumn{7}{c|}{Enhancing Tumor}                     & \multicolumn{7}{c|}{Tumor Core}                          & \multicolumn{7}{c|}{Whole Tumor}                         \\ \hline
Fl         & T1        & T1c      & T2        & HMIS & HVED & RbSeg & mmFm & ShSpc  & MetaKD & Imprv & HMIS & HVED & RbSeg & mmFm & ShSpc  & MetaKD & Imprv & HMIS & HVED & RbSeg & mmFm & ShSpc  & MetaKD & Imprv \\ \hline
$\bullet$ & $\circ$   & $\circ$   & $\circ$   & 11.78   & 23.80   & 25.69     & 39.33    & \textit{43.52} & \textbf{47.37} & \greent{8.85\%} & 26.06   & 57.90   & 53.57     & 61.21    & \textit{69.44} & \textbf{73.01} & \greent{5.14\%} & 52.48   & 84.39  & 85.69     & 86.10     & \textit{88.68} & \textbf{89.11} & \greent{0.48\%} \\
$\circ$   & $\bullet$ & $\circ$   & $\circ$   & 10.16   & 8.60    & 17.29     & 32.53    & \textit{41.00}    & \textbf{44.94} & \greent{9.60\%} & 37.39   & 33.90   & 47.90      & 56.55    & \textit{63.18} & \textbf{67.16} & \greent{6.30\%} & 57.62   & 49.51  & 70.11     & 67.52    & \textit{73.44} & \textbf{77.47} & \greent{5.49\%} \\
$\circ$   & $\circ$   & $\bullet$ & $\circ$   & 62.02   & 57.64  & 67.07     & 72.60     & \textit{73.29} & \textbf{76.22} & \greent{4.00\%}  & 65.29   & 59.59  & 76.83     & 75.41    & \textit{78.65} & \textbf{81.98} & \greent{4.12\%}  & 61.53   & 53.62  & 73.31     & 72.22    & \textit{73.82} & \textbf{77.72} & \greent{5.28\%} \\
$\circ$   & $\circ$   & $\circ$   & $\bullet$ & 25.63   & 22.82  & 28.97     & 43.05    & \textit{46.31} & \textbf{46.93} & \greent{1.34\%} & 57.20    & 54.67  & 57.49     & 64.20     & \textit{69.03} & \textbf{69.72} & \greent{1.00\%} & 80.96   & 79.83  & 82.24     & 81.15    & \textit{83.99} & \textbf{84.44} & \greent{0.54\%} \\
$\bullet$ & $\bullet$ & $\circ$   & $\circ$   & 10.71   & 27.96  & 32.13     & 42.96    & \textit{44.76} & \textbf{49.64} & \greent{10.90\%} & 41.12   & 61.14  & 60.68     & 65.91    & \textit{72.67} & \textbf{75.66} & \greent{4.11\%} & 64.62   & 85.71  & 88.24     & 87.06    & \textit{89.76} & \textbf{90.23} & \greent{0.52\%} \\
$\bullet$ & $\circ$   & $\bullet$ & $\circ$   & 66.10   & 68.36  & 70.30     & 75.07    & \textit{75.60} & \textbf{77.49} & \greent{2.50\%} & 71.49   & 75.07  & 80.62     & 77.88    & \textit{84.50} & \textbf{84.82} & \greent{0.38\%} & 68.99   & 85.93  & 88.51     & 87.30     & \textit{90.06} & \textbf{90.30} & \greent{0.27\%} \\
$\bullet$ & $\circ$   & $\circ$   & $\bullet$ & 30.22   & 32.31  & 33.84     & \textit{47.52}    & 47.20 & \textbf{50.27} & \greent{5.79\%} & 57.68   & 62.70  & 61.16     & 69.75    & \textit{72.93} & \textbf{75.88} & \greent{4.04\%} & 82.95   & 87.58  & 88.28     & 87.59    & \textit{90.02} & \textbf{90.39} & \greent{0.41\%} \\
$\circ$   & $\bullet$ & $\bullet$ & $\circ$   & 66.22   & 61.11  & 69.06     & 74.04    & \textbf{75.76} & \textit{75.41} & \redt{-0.46\%} & 72.46   & 67.55  & 78.72     & 78.59    & \textbf{82.10} & \textit{81.85} & \redt{-0.30\%} & 68.47   & 64.22  & 77.18     & 74.42    & \textit{78.74} & \textbf{80.59} & \greent{2.35\%} \\
$\circ$   & $\bullet$ & $\circ$   & $\bullet$ & 32.39   & 24.29  & 32.01     & 44.99    & \textit{46.84} & \textbf{49.77} & \greent{4.12\%} & 60.92   & 56.26  & 62.19     & 69.42    & \textit{71.38} & \textbf{74.40} & \greent{4.23\%} & 82.41   & 81.56  & 84.78     & 82.20    & \textit{86.03} & \textbf{87.05} & \greent{1.19\%} \\
$\circ$   & $\circ$   & $\bullet$ & $\bullet$ & 67.83   & 67.83  & 69.71     & 74.51    & \textit{75.95} & \textbf{78.24} & \greent{3.02\%} & 76.64   & 73.92  & 80.20     & 78.61    & \textit{83.82} & \textbf{84.84} & \greent{1.22\%} & 82.48   & 81.32  & 85.19     & 82.99    & \textit{85.42} & \textbf{86.18} & \greent{0.89\%} \\
$\bullet$ & $\bullet$ & $\bullet$ & $\circ$   & 68.54   & 68.60  & 70.78     & 75.47    & \textit{76.42} & \textbf{77.66} & \greent{1.62\%} & 76.01   & 77.05  & 81.06     & 79.80    & \textit{85.23} & \textbf{85.36} & \greent{0.15\%} & 72.31   & 86.72  & 88.73     & 87.33    & \textit{90.29} & \textbf{90.70} & \greent{0.45\%} \\
$\bullet$ & $\bullet$ & $\circ$   & $\bullet$ & 31.07   & 32.34  & 36.41     & \textit{47.70}    & 46.55 & \textbf{51.05} & \greent{7.02\%} & 60.32   & 63.14  & 64.38     & 71.52    & \textit{73.97} & \textbf{76.73} & \greent{3.73\%} & 83.43   & 88.07  & 88.81     & 87.75    & \textit{90.36} & \textbf{90.59} & \greent{0.25\%} \\
$\bullet$ & $\circ$   & $\bullet$ & $\bullet$ & 68.72   & 68.93  & 70.88     & \textit{75.67}    & \textbf{75.99} & 75.58 & \redt{-0.54\%} & 77.53   & 76.75  & 80.72     & 79.55    & \textbf{85.26} & \textit{85.08} & \redt{-0.21\%} & 83.85   & 88.09  & 89.27     & 88.14    & \textit{90.78} & \textbf{90.87} & \greent{0.10\%} \\
$\circ$   & $\bullet$ & $\bullet$ & $\bullet$ & 69.92   & 67.75  & 70.10     & 74.75    & \textbf{76.37} & \textit{75.87} & \redt{-0.65\%} & 78.96   & 75.28  & 80.33     & 80.39    & \textit{84.18} & \textbf{84.57} & \greent{0.46\%} & 83.94   & 82.32  & 86.01     & 82.71    & \textit{86.47} & \textbf{86.94} & \greent{0.54\%} \\
$\bullet$ & $\bullet$ & $\bullet$ & $\bullet$ & 70.24   & 69.03  & 71.13     & 77.61    & \textit{78.08} & \textbf{79.74} & \greent{2.13\%} & 79.48   & 77.71  & 80.86     & \textit{85.78}    & 85.45 & \textbf{86.20} & \greent{0.49\%} & 84.74   & 88.46  & 89.45     & 89.64    & \textit{90.88} & \textbf{90.92} & \greent{0.04\%} \\ \hline
\multicolumn{4}{|c|}{Average}                   & 46.10   & 46.76  & 51.02     & 59.85    & \textit{61.58} & \textbf{63.74} & \greent{3.51\%} & 62.57   & 64.84  & 69.78     & 72.97    & \textit{77.45} & \textbf{79.15} & \greent{2.19\%} & 74.05   & 79.16  & 84.39     & 82.94    & \textit{85.92} & \textbf{86.90} & \greent{1.14\%} \\ \hline
\multicolumn{4}{|c|}{p-value}                   & 3.5e-6   & 2.2e-7  & 7.3e-7     & 6.2e-5    & 7.3e-5 & N/A & N/A & 1.6e-5   & 9.7e-8  & 4.2e-6     & 1.0e-7    & 1.8e-4 & N/A & N/A & 3.8e-5   & 6.7e-4  & 6.3e-6     & 4.5e-7    & 3.2e-3 & N/A & N/A \\ \hline
\end{tabular}
}\end{center}
\end{table*}
\begin{table*}[h]
\scriptsize
\setlength\tabcolsep{1pt}
\begin{center}
\caption{Model performance comparison on the test set for the ADNI \textbf{classification} task across all possible modality combinations. The table reports classification accuracy (in \%)  and F1-Score (normalized into 0-100) for predicting Alzheimer's disease progression using multi-modal inputs (fMRI, genomic, clinical, and biospecimen data). Our proposed MetaKD model is compared with LiMoE \cite{mustafa2022multimodal}, Flex-MOE \cite{yun2024flex}, and MoE-Retriever \cite{yun2025generate}.% \hu{Balanced Accuracy, micro F1 score. imbalance and any tweaking in losses. explain more}
}
\label{tab:adni}
\resizebox{1.0\linewidth}{!}{
\begin{tabular}{|cccc|ccccc|ccccc|ccccc|ccccc|} 
\hline
\multicolumn{4}{|c|}{Modalities} & \multicolumn{5}{c|}{Accuracy} & \multicolumn{5}{c|}{F1-Score} & \multicolumn{5}{c|}{Balanced Accuracy} & \multicolumn{5}{c|}{micro F1-Score}\\
\hline
C & G & B & I & LiMoE & Flex-MoE & MoE-Ret & MetaKD & Imprv & LiMoE & Flex-MoE & MoE-Ret & MetaKD & Imprv & LiMoE & Flex-MoE & MoE-Ret & MetaKD & Imprv & LiMoE & Flex-MoE & MoE-Ret & MetaKD & Imprv\\
\hline
$\bullet$ & $\circ$ & $\circ$ & $\circ$ & 53.04 & 54.94 & \textit{57.12} & \textbf{59.65} & \greent{4.43\%} & 40.56 & 42.52 & \textit{42.89} & \textbf{43.25} & \greent{0.84\%} & 48.32 & 50.15 & \textit{53.78} & \textbf{56.24} & \greent{4.58\%} & 52.87 & 54.78 & \textit{56.91} & \textbf{59.29} & \greent{4.18\%}\\
$\circ$ & $\bullet$ & $\circ$ & $\circ$ & 56.40 & \textbf{78.21} & \textit{76.85} & 77.32 & \redt{-1.14\%} & 42.90 & 46.16 & \textit{46.82} & \textbf{47.32} & \greent{1.07\%} & 51.23 & \textbf{72.45} & \textit{71.34} & 71.89 & \redt{-0.77\%} & 56.12 & \textbf{77.89} & \textit{76.53} & 77.01 & \redt{-1.13\%}\\
$\circ$ & $\circ$ & $\bullet$ & $\circ$ & 46.65 & \textbf{58.31} & \textit{57.24} & 56.23 & \redt{-3.56\%} & 32.64 & 36.51 & \textit{38.95} & \textbf{40.57} & \greent{4.16\%} & 42.87 & \textbf{55.46} & \textit{54.12} & 53.46 & \redt{-3.60\%} & 46.34 & \textbf{57.94} & \textit{56.78} & 55.98 & \redt{-3.38\%}\\
$\circ$ & $\circ$ & $\circ$ & $\bullet$ & 55.85 & \textbf{56.38} & \textit{55.89} & 55.32 & \redt{-1.88\%} & 36.78 & 38.74 & \textit{38.91} & \textbf{39.05} & \greent{0.36\%} & 51.42 & \textbf{52.16} & \textit{52.05} & 51.89 & \redt{-0.52\%} & 55.67 & \textbf{56.21} & \textit{55.73} & 55.14 & \redt{-1.90\%}\\
$\bullet$ & $\bullet$ & $\circ$ & $\circ$ & - & 54.12 & \textit{58.67} & \textbf{61.35} & \greent{4.57\%} & - & 31.89 & \textit{40.12} & \textbf{42.36} & \greent{5.58\%} & - & 49.87 & \textit{55.23} & \textbf{57.92} & \greent{4.87\%} & - & 53.89 & \textit{58.41} & \textbf{61.08} & \greent{4.57\%}\\
$\bullet$ & $\circ$ & $\bullet$ & $\circ$ & - & 57.98 & \textit{60.45} & \textbf{63.24} & \greent{4.61\%} & - & 35.48 & \textit{43.21} & \textbf{45.87} & \greent{6.15\%} & - & 53.65 & \textit{57.12} & \textbf{59.78} & \greent{4.66\%} & - & 57.65 & \textit{60.18} & \textbf{62.95} & \greent{4.60\%}\\
$\bullet$ & $\circ$ & $\circ$ & $\bullet$ & 56.08 & 57.24 & \textit{59.87} & \textbf{62.75} & \greent{4.81\%} & 38.45 & 36.25 & \textit{39.64} & \textbf{41.23} & \greent{4.02\%} & 52.34 & 53.78 & \textit{56.95} & \textbf{59.42} & \greent{4.33\%} & 55.89 & 57.06 & \textit{59.67} & \textbf{62.51} & \greent{4.76\%}\\
$\circ$ & $\bullet$ & $\bullet$ & $\circ$ & - & 56.43 & \textit{60.89} & \textbf{63.76} & \greent{4.71\%} & - & 30.56 & \textit{36.78} & \textbf{38.63} & \greent{5.03\%} & - & 52.18 & \textit{57.67} & \textbf{60.34} & \greent{4.63\%} & - & 56.21 & \textit{60.67} & \textbf{63.52} & \greent{4.69\%}\\
$\circ$ & $\bullet$ & $\circ$ & $\bullet$ & 56.98 & 59.25 & \textit{62.13} & \textbf{64.87} & \greent{4.41\%} & 40.87 & 37.25 & \textit{42.56} & \textbf{44.21} & \greent{3.88\%} & 53.45 & 55.89 & \textit{59.21} & \textbf{61.78} & \greent{4.34\%} & 56.76 & 59.03 & \textit{61.91} & \textbf{64.65} & \greent{4.42\%}\\
$\circ$ & $\circ$ & $\bullet$ & $\bullet$ & 48.16 & 51.58 & \textit{54.78} & \textbf{57.56} & \greent{5.08\%} & 33.12 & 34.58 & \textit{37.23} & \textbf{39.54} & \greent{6.21\%} & 44.89 & 48.34 & \textit{52.11} & \textbf{54.67} & \greent{4.91\%} & 47.98 & 51.36 & \textit{54.56} & \textbf{57.32} & \greent{5.07\%}\\
$\bullet$ & $\bullet$ & $\bullet$ & $\circ$ & - & 55.82 & \textit{57.89} & \textbf{60.35} & \greent{4.25\%} & - & 38.74 & \textit{40.67} & \textbf{42.11} & \greent{3.54\%} & - & 51.96 & \textit{54.78} & \textbf{57.23} & \greent{4.47\%} & - & 55.61 & \textit{57.67} & \textbf{60.12} & \greent{4.25\%}\\
$\bullet$ & $\bullet$ & $\circ$ & $\bullet$ & - & 60.54 & \textit{62.34} & \textbf{65.04} & \greent{4.33\%} & - & 48.67 & \textit{49.23} & \textbf{50.43} & \greent{2.44\%} & - & 57.38 & \textit{59.67} & \textbf{62.15} & \greent{4.16\%} & - & 60.32 & \textit{62.12} & \textbf{64.82} & \greent{4.35\%}\\
$\bullet$ & $\circ$ & $\bullet$ & $\bullet$ & - & 53.35 & \textit{59.78} & \textbf{63.06} & \greent{5.49\%} & - & \textbf{43.21} & \textit{41.89} & 42.65 & \redt{-1.30\%} & - & 49.78 & \textit{56.67} & \textbf{59.89} & \greent{5.67\%} & - & 53.12 & \textit{59.56} & \textbf{62.84} & \greent{5.51\%}\\
$\circ$ & $\bullet$ & $\bullet$ & $\bullet$ & - & 61.23 & \textit{62.45} & \textbf{64.12} & \greent{2.67\%} & - & 52.32 & \textit{53.78} & \textbf{55.09} & \greent{2.44\%} & - & 58.45 & \textit{60.23} & \textbf{61.78} & \greent{2.57\%} & - & 61.01 & \textit{62.23} & \textbf{63.89} & \greent{2.67\%}\\
$\bullet$ & $\bullet$ & $\bullet$ & $\bullet$ & - & 65.23 & \textit{66.45} & \textbf{67.87} & \greent{2.18\%} & - & 53.45 & \textit{55.67} & \textbf{57.23} & \greent{2.82\%} & - & 62.78 & \textit{64.12} & \textbf{65.34} & \greent{1.90\%} & - & 65.01 & \textit{66.23} & \textbf{67.65} & \greent{2.14\%}\\
\hline
\multicolumn{4}{|c|}{Average} & 53.31 & 58.71 & \textit{60.78} & \textbf{62.83} & \greent{3.37\%} & 37.90 & 40.42 & \textit{42.89} & \textbf{44.64} & \greent{4.08\%} & 49.86 & 55.32 & \textit{57.89} & \textbf{59.98} & \greent{3.61\%} & 53.09 & 58.48 & \textit{60.56} & \textbf{62.60} & \greent{3.37\%}\\
\hline
\multicolumn{4}{|c|}{p-value} & 5.8e-3 & 5.2e-5 & 2.1e-4 & N/A & N/A & 9.5e-4 & 1.7e-4 & 3.2e-4 & N/A & N/A & 4.2e-3 & 3.8e-5 & 1.9e-4 & N/A & N/A & 6.1e-3 & 4.9e-5 & 2.3e-4 & N/A & N/A\\
\hline
\end{tabular}
}
\end{center}
\end{table*}
Table \ref{tab:brats2018} presents the overall performance for all 15 possible combinations of missing modalities during testing for segmenting the three sub-regions of brain tumors from BraTS2018.\footnote{We only compare our model with ``non-dedicated'' strategies. ``Dedicated'' strategies, e.g. KD-Net~\cite{hu2020knowledge} and ACN~\cite{wang2021acn}, are not considered to be compared here. The comparison with RFNet~\cite{ding2021rfnet} is in the Appendix.}
We compare our models with several competing models, including U-HeMIS (abbreviated as HMIS in the table)~\cite{havaei2016hemis}, U-HVED (HVED)~\cite{dorent2019hetero}, Robust-MSeg (RSeg)~\cite{chen2019robust}, mmFormer (mmFm)~\cite{zhang2022mmformer} and ShaSpec (ShSpc)~\cite{wang2023multi}. The performance improvements between MetaKD and the second-best model are also shown in the table by calculating the gaps between the two and divided by the compared model performance. Notably, when T1c is available, all models exhibit superior performance for the enhancing tumor (ET) segmentation compared to other modalities. Similarly, T1c for the tumor core (TC) and Flair for the whole tumor (WT) show superior results compared to other modalities, thereby validating our motivation for distilling the knowledge from the most informative modalities in multi-modal learning.

Our MetaKD model demonstrates significant superiority over U-HeMIS, U-HVED, Robust-MSeg, mmFormer and ShaSpec in terms of segmentation Dice for whole tumor in all 15 combinations. For enhancing tumor and tumor core, it shows the best performance in 12 and 13 out of 15 combinations, respectively. 
Using the average results for the models, we compute the one-tailed paired t-test results for each sub-task, as shown in the last row of Table \ref{tab:brats2018}, and the results confirm the superiority of MetaKD. On average, the proposed MetaKD model outperforms the state-of-the-art performance by 3.51\% for enhancing tumor, 2.19\% for tumor core, and 1.14\% for whole tumor in terms of the segmentation Dice score.

It is worth noting that in some combinantions where the best modality is missing, such as ET/TC without T1c and WT without Flair, the MetaKD model exhibits a remarkable performance improvement. For instance, compared with the second best model, the ET segmentation shows an 8.85\% improvement when using only the Flair modality; 9.60\% improvement when using only the T1 modality; and a striking 10.90\% improvement when using Flair and T1 modalities. 
For TC segmentation, it achieves 5.14\% and 6.30\% improvement with respect to the second-best model when using only Flair and T1, respectively. Similarly, for WT segmentation, it achieves 5.49\% and 5.28\% improvement with respect to the second-best model when using only T1 and T1c, respectively. 
These outstanding results confirm that the MetaKD model can distill valuable knowledge from the best modalities when training the model for multi-modal learning with missing modalities.

\subsection{Model Performance on Medical Image Classification}

Table \ref{tab:adni} presents a comprehensive performance evaluation of our proposed MetaKD model against established baselines, including LiMoE~\cite{mustafa2022multimodal}\footnote{LiMoE is designed to handle only one tabular, one imaging, or a combination of one each at a time due to its expert routing mechanism, which assigns each input to specialized experts. Multiple tabular modality inputs are not supported without further architectural modifications.} The current state-of-the-art Flex-MOE~\cite{yun2024flex} and MoE-Retriever, which employs a mixture-of-experts architecture with retrieval-augmented mechanisms to enhance multi-modal fusion through external knowledge integration, on the ADNI multi-class classification task. This task involves distinguishing between three classes: control (CTL), mild cognitive impairment (MCI), and AD patients. This categorization is crucial for early detection and monitoring of decline progression. We evaluate the results using the accuracy and F1-score to provide a comprehensive view of the model's performance across all possible modality combinations, including Clinical (C), Genomic (G), Biospecimen (B), and Imaging (I) data. Overall, MetaKD demonstrates superior performance, achieving an average accuracy of 62.83\% compared to Flex-MOE's 58.71\% and an average F1-score of 44.64 versus 40.42. From modality performance, we can see genomic data is the most important modality, while clinical data shows a notable improvement with MetaKD (59.65\% v.s. 54.94\%).

The most clinically significant findings emerge in multi-modal scenarios, where MetaKD's meta-learning framework demonstrates superior capability in handling missing modalities—a common challenge in clinical practice due to patient compliance, cost considerations, and data acquisition limitations. The Clinical + Biospecimen + Imaging combination yields an 18.20\% accuracy improvement (63.06\% vs. 54.12\%), suggesting that MetaKD can effectively compensate for missing genomic data, which is often unavailable in routine clinical settings.

Particularly noteworthy is the Clinical + Genomic combination's performance (61.35\% accuracy, 13.36\% improvement), as this represents a practical scenario where expensive imaging studies might be contraindicated or unavailable. The 32.86\% F1-score improvement (42.36 vs. 31.89) in this combination indicates enhanced discrimination across all three diagnostic categories, which is crucial for accurate staging of cognitive decline.

These results suggest that MetaKD effectively leverages complementary information from multiple modalities. F1-score improvements are particularly notable in certain combinations, with Clinical + Genomic showing a 32.86\% improvement (42.36 v.s. 31.89) and Clinical + Biospecimen demonstrating a 29.29\% improvement (45.87 v.s. 35.48). 
These results validate MetaKD's effectiveness in handling missing modalities and leveraging cross-modal information for improved Alzheimer's disease classification, particularly in scenarios where multiple, but not all, modalities are available.

\subsection{Model Performance on General Image Classification}

In line with the SMIL setup~\cite{ma2021smil}, we conduct the training of MetaKD on both partial and full modality sub-datasets, comprising images and audios. Specifically in Tab.~\ref{tab:avmnist}, we create the missing modality sub-dataset by setting the audio modality rates to $\{ 5\%, 10\%, 15\%, 20\% \}$, which determines the proportion of available audio data used for training. In this missing audio setup, the visual modality data is fully available for training and testing.
Similarly, in Tab.~\ref{tab:avmnist-missImg}, we set the visual modality rates to $\{ 5\%, 10\%, 15\%, 20\% \}$, which is the percentage of available visual data for training. In this missing visual data setup, the audio modality data is fully available for training and testing.

\begin{table*}[h]
  \caption{Model performance comparison of testing classification accuracy (in \%) of missing audio (\ref{tab:avmnist}) and visual (\ref{tab:avmnist-missImg}) modalities (by setting different available audio and visual rates) on Audiovision-MNIST dataset. The lower bound (LowerB) is a LeNet~\cite{lecun1998gradient} network trained with single modality, i.e., images only in (\ref{tab:avmnist}) and audio only in (\ref{tab:avmnist-missImg}). The upper bound (UpperB) is a model trained with all data modalities (all images and audios). 
The best results for each row are in bold.}
  \label{tab:main_table}
  \centering
  \begin{subtable}{0.6\linewidth}
    \centering
    \scalebox{0.6}{
    \begin{tabular}{|c|cc|ccccc|c|}
\hline
Audio rate & LowerB & UpperB & AutoEncoder & GAN   & Full2miss & SMIL  & ShaSpec & MetaKD           \\ \hline
5\%        & 92.35  & 98.22  & 89.78       & 89.11 & 90.00     & 92.89 & 93.33   & \textbf{93.56} \\
10\%       & 92.35  & 98.22  & 89.33       & 89.78 & 91.11     & 93.11 & 93.56   & \textbf{94.22} \\
15\%       & 92.35  & 98.22  & 89.78       & 88.67 & 92.23     & 93.33 & 93.78   & \textbf{94.89} \\
20\%       & 92.35  & 98.22  & 88.89       & 89.56 & 92.67     & 94.44 & 94.67   & \textbf{95.11} \\ \hline
\end{tabular}
}
    \caption{Missing audio data.}
    \label{tab:avmnist}
  \end{subtable}%
  \hfill
  \begin{subtable}{0.4\linewidth}
    \centering
    \scalebox{0.6}{
    \begin{tabular}{|c|cc|cc|c|}
\hline
Visual rate & LowerB & UpperB & SMIL  & ShaSpec & MetaKD  \\ \hline
5\%        & 80.44  & 98.22  & 83.33 & 84.00   & \textbf{84.67} \\
10\%       & 80.44  & 98.22  & 84.89 & 85.11   & \textbf{85.56} \\
15\%       & 80.44  & 98.22  & 85.56 & 86.67   & \textbf{86.89} \\
20\%       & 80.44  & 98.22  & 87.78 & 88.22   & \textbf{88.44} \\ \hline
\end{tabular}
}
    \caption{Missing visual data.}
    \label{tab:avmnist-missImg}
  \end{subtable}
\end{table*}

During the evaluation phase, only images are provided as input to the models in Tab.~\ref{tab:avmnist}, and only audio data is provided as input to the models in Tab.~\ref{tab:avmnist-missImg}.
To assess the performance of MetaKD, we compare it against several other models, including Auto-encoder~\cite{baldi2012autoencoders}, a Generative Adversarial Network (GAN) based model~\cite{goodfellow2020generative}, a method for distilling multi-modality knowledge to train missing modality models~\cite{shen2019brain}, SMIL~\cite{ma2021smil} and ShaSpec~\cite{wang2023multi}. 
We set a LeNet~\cite{lecun1998gradient} network with single modality (images only) as the lower bound for testing, while a model trained with full modalities (including all images and audios) serves as the upper bound for testing.
For the missing visual data experiment, we only compare MetaKD with SMIL and ShaSpec because they are the most accurate models and they also have code available.

As depicted in Tab.~\ref{tab:avmnist}, MetaKD exhibits strong performance, achieving an accuracy of 94.22\% compared to the second-best model with 93.56\% at an audio rate of 10\%, and 94.89\% vs. 93.78\% at an audio rate of 15\%. Notably, as the audio rate increases, the performance of all models improves, with MetaKD consistently outperforming all other models.
Similarly, in Tab.~\ref{tab:avmnist-missImg}, MetaKD improves by around $0.5\%$ at visual rates of 5\% and 10\% compared to the second-best models, and with increasing visual rate, accuracy also improves for all models, where MetaKD is consistently better.
This can be attributed to MetaKD's exceptional ability to retain information-rich representations from important modalities and effectively use the available features to compensate for the missing modalities. 
Another interesting observation is that the gains from $\approx$1\% to $\approx$3\% in Tab.~\ref{tab:avmnist} from LowerB (model trained with visual data only) are much smaller than the gains of $\approx$4\% to $\approx$8\% in Tab.~\ref{tab:avmnist-missImg} from LowerB (model trained with audio data only). This suggests that the visual data is more important than the audio data because the improvements in Tab.~\ref{tab:avmnist-missImg} (when audio data is distilling the knowledge from the missing visual data) are more remarkable than the improvements in Tab.~\ref{tab:avmnist} (when visual data is distilling the knowledge from the missing audio data).

\subsection{Analyses of MetaKD}

In this section, we first provide a sensitivity analysis of $\alpha$ in Eq.~\eqref{eq:meta} and the different loss functions for $\ell_{kd}(.)$ in Eq.~\eqref{eq:kd}. 
We then show how the importance of weight vectors (IWV) varies during training.
We also provide visualization results of multi-modal segmentation (on BraTS2018) with missing modalities, where we compare the results with baseline methods consisting of MetaKD without the modality-weighted knowledge distillation (modality-weighted knowledge distillation) from Eq.~\eqref{eq:kd}, and results by ShaSpec~\cite{wang2023multi} and the proposed MetaKD.
We conclude the section with a study on the effectiveness of our generated features, as explained in Eq.~\eqref{eq:ftgen}.

\begin{figure}[]
\centering
\includegraphics[width=0.3\textwidth]{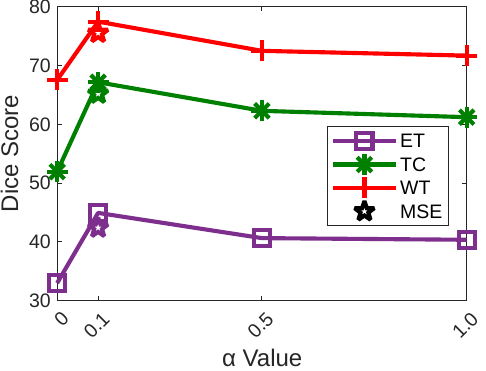}
\caption{We show Dice as a function of $\alpha$ on the BraTS2018 dataset with T1 modality only for testing, representing the L1 loss in Eq.~\ref{eq:kd} for $\ell_{kd}(.)$. The Dice scores obtained using the L2 loss for $\ell_{kd}(.)$ ($\alpha = 0.1$) are marked with star symbols (MSE). Different tumor regions are represented by distinct colors.
}
\label{fig:sen}
\end{figure}

\begin{figure}[]
\centering
\includegraphics[width=0.43\textwidth]{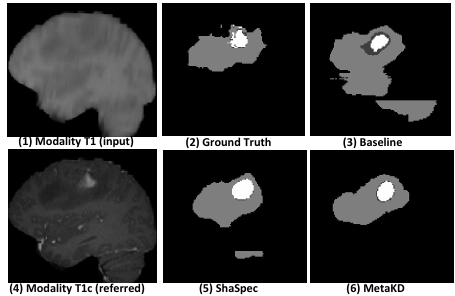}
\caption{We show the segmentation results produced by different models using T1 only as input for testing on BraTS2018. The images in (1) and (4) represent the MRI T1 (input data for the model) and T1c modalities (referred data, not used by the model), respectively. 
The  ``Baseline'' model (3) refers to MetaKD without the modality-weighted knowledge distillation process, while the results for ShaSpec~\cite{wang2023multi} and our MetaKD are shown in (5) and (6). White, dark gray and light gray correspond to distinct ET, TC and WT tumor sub-regions, respectively.
}
\label{fig:vis-seg}
\end{figure}

\noindent\textbf{Sensitivity Analysis of $\alpha$ and Different Loss Functions for $\ell_{kd}(.)$.} In Fig.~\ref{fig:sen}, 
we vary the value of $\alpha$ of Eq.~\eqref{eq:meta}, utilizing solely the T1 input for testing and applying the L1 loss for $\ell_{kd}(.)$ in Eq.~\eqref{eq:kd}. 
Notably, a dramatic drop in model performance is observed when $\alpha$ equals 0. However, positive values of $\alpha$ lead to performance improvements, with the optimal result achieved at $\alpha = 0.1$.
To investigate the impact of a different p-norm for the modality-weighted knowledge distillation loss $\ell_{kd}(.)$ in Eq.~\eqref{eq:kd}, we evaluate the Dice score using the L2 distance as the metric (i.e., MSE loss), while keeping $\alpha$ fixed at 0.1. A comparison between the L1 and MSE losses reveals a slight reduction in Dice score with the MSE loss over three types of tumors, showing that the L1 loss is a better option.

\noindent\textbf{Qualitative Visualization.} Figure~\ref{fig:vis-seg} shows segmentation visualizations for different models on the BraTS2018 dataset. The Baseline model (MetaKD without modality-weighted knowledge distillation) exhibits more errors compared to ShaSpec~\cite{wang2023multi} and MetaKD. Although both models have room for improvement with the T1-only modality, MetaKD achieves better results, particularly at tumor boundaries, due to its ability to retain valuable cross-modal knowledge from important modalities (e.g., T1c) even when only T1 is available.

\begin{figure*}[]
\centering
\begin{subfigure}{.32\linewidth}
\includegraphics[width=1.0\linewidth]{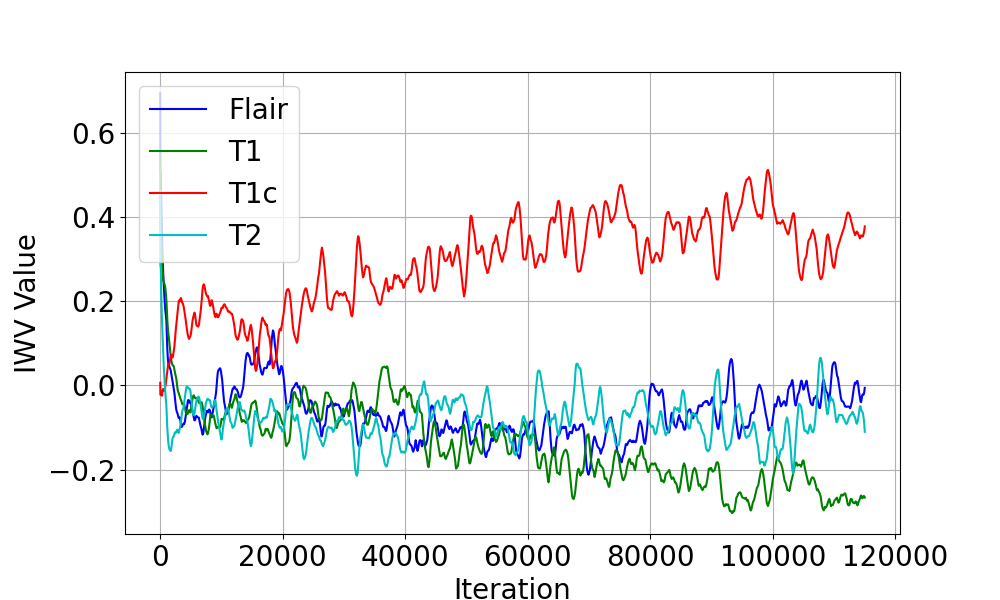}
\caption{IWV values on BraTS2018.}\label{fig:iwv}
\end{subfigure}
\begin{subfigure}{.32\linewidth}
\includegraphics[width=\linewidth]{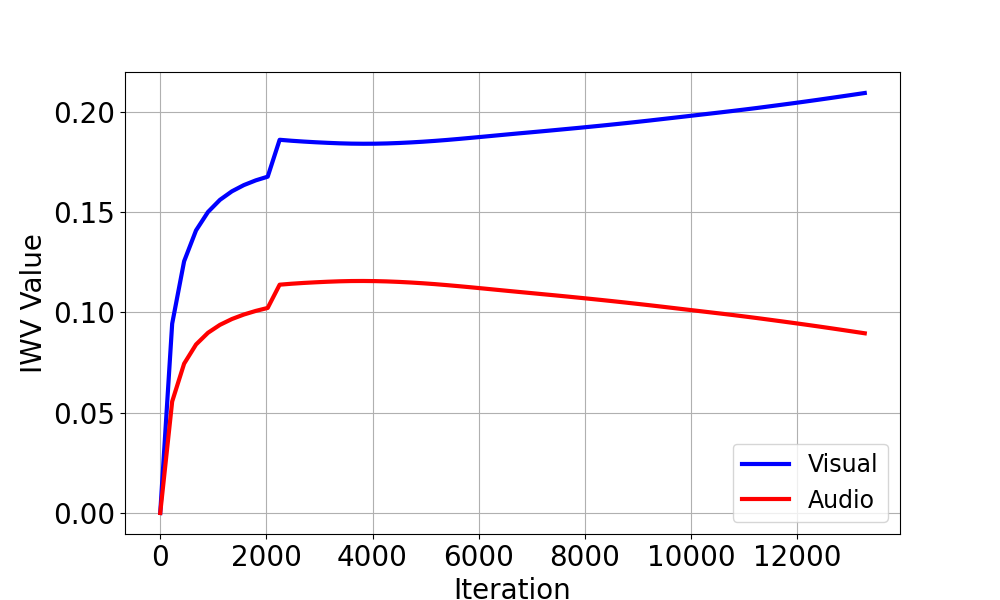}
\caption{IWV values on AV-MNIST.}\label{fig:avmnist_iwv}
\end{subfigure}
\begin{subfigure}{.32\linewidth}
\includegraphics[width=\linewidth]{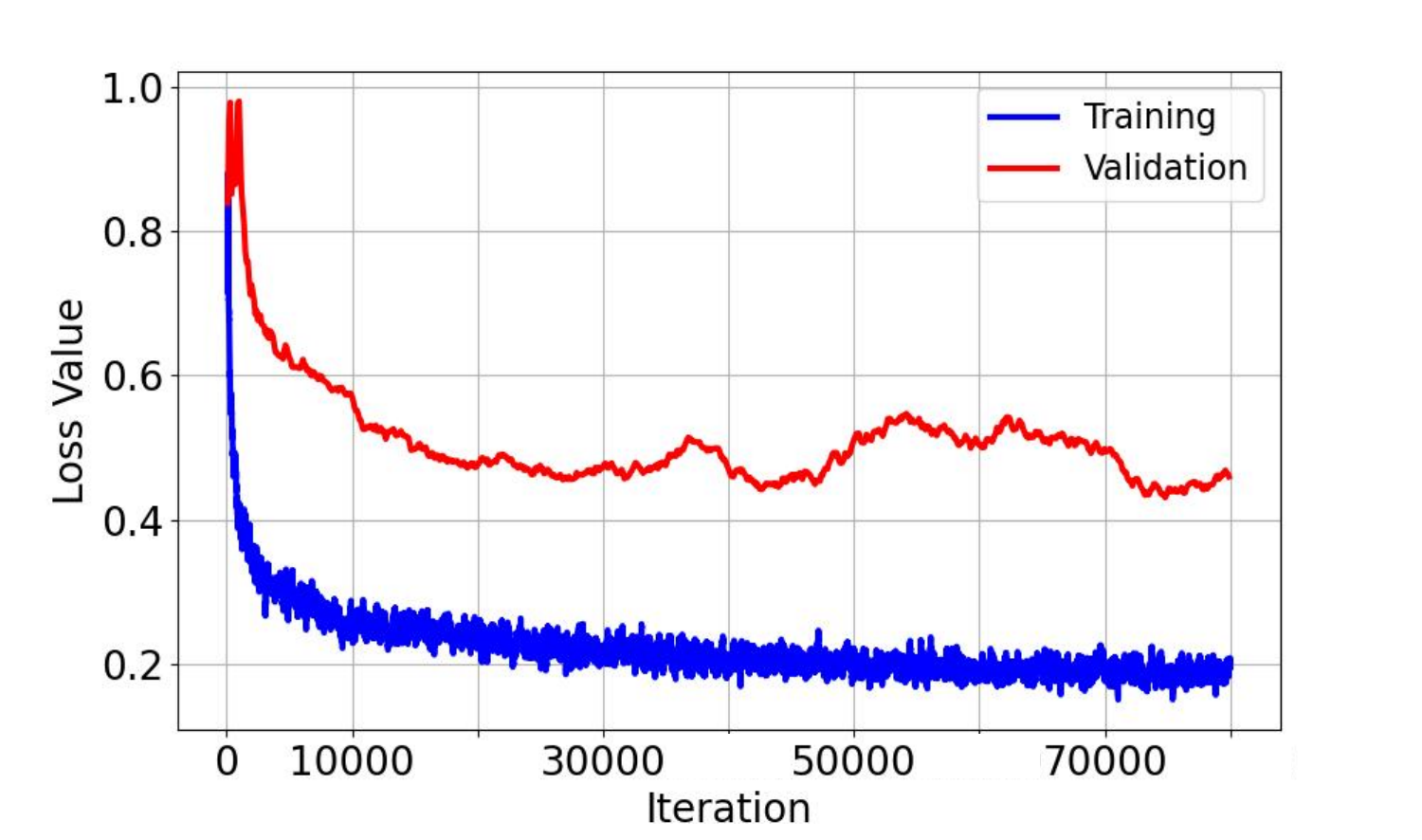}
\caption{Train/Val losses.}\label{fig:trainval_loss}
\end{subfigure}
\caption{Modality IWV weights changes over training iterations on BraTS2018 dataset (a) and AudioVision-MNIST (b). In (c), we have the training and validation segmentation losses on BraTS2018. The IWV values for ADNI can be found in supplementary material.
}
\label{fig:weight_train_val_loss}
\end{figure*}

\noindent\textbf{The Optimization of IWV.} The progress of IWV during training is displayed in Figures~\ref{fig:iwv} and~\ref{fig:avmnist_iwv}. The optimization and changes of IWV weights corresponding to different modalities 
are denoted using curves with different colors.
By normalizing the IWV weight vector with a softmax activation function, as explained in Eq.~\eqref{eq:IWV_norm}, we guarantee that the sum of each element within the IWV weight vector equals to 1, which ensures that the relative importance of the modalities are kept, where changes in the weight of one element will affect the weights in other elements. In our training, at least one modality will dominate the others and consequently receive a high IWV weight.
This large IWV weight modality will then be used by other modalities in a knowledge distillation process. 
As shown in  Figure~\ref{fig:iwv}, the IWV weight of T1c (in red) is above other curves, showing the significance of the modality T1c for Brain Tumor detection (particularly for enhancing tumor and tumor core). 
This observation resonates with the fact that enhancing tumor and tumor core is clearly visible in T1c~\cite{chen2019robust}. 
Interestingly, we also notice that the IWV weight of Flair (in dark blue) is smaller than the weight for T1c, but it has a sharp increase at around 10000 iterations, suggesting the importance of this modality at that training stage. After that, the Flair weight decreases and becomes similar to the weights of T1 and T2, but still slightly higher than them at the end of the training. 
This phenomenon may be attributed to the effective distillation of Flair features at the beginning of the optimization (particularly for the whole tumor); then the model subsequently prioritizes the distillation of T1c features. This pattern also holds for AudioVision classification shown by Fig.~\ref{fig:avmnist_iwv}. It shows the visual modality receives
higher importance gradually than the audio modality. 
An important question is if our training dynamics lead to the overfitting of the dominant modality.
However, the training and validation losses shown in Fig.~\ref{fig:trainval_loss} suggest that such training dynamics do not result in overfitting.

\begin{figure}[]
\centering
\begin{subfigure}{.49\linewidth}
\includegraphics[width=1.0\linewidth]{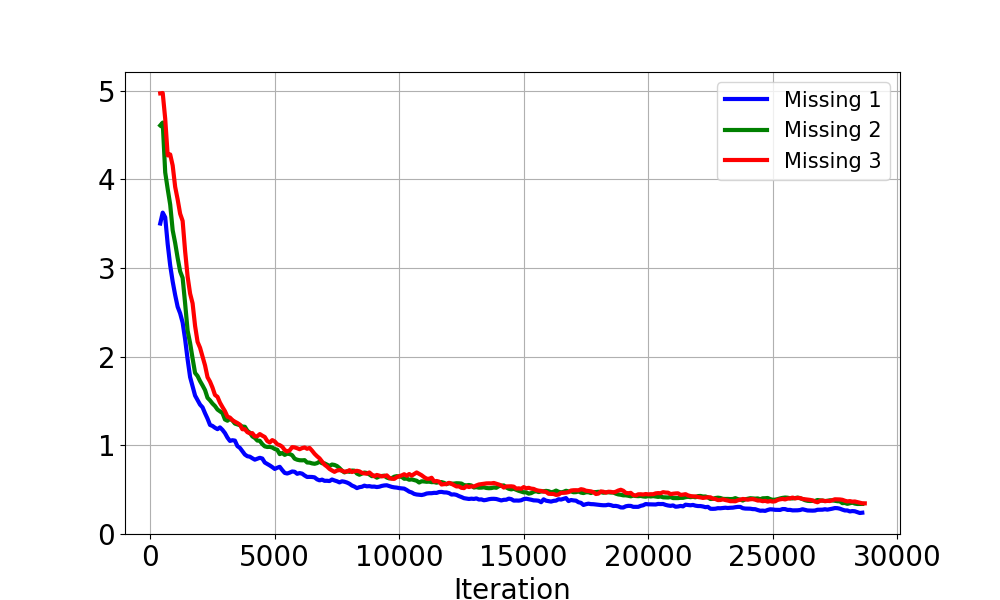}
\caption{L2 distance.}\label{fig:ft_dist}
\end{subfigure}
\begin{subfigure}{.49\linewidth}
\includegraphics[width=1.0\linewidth]{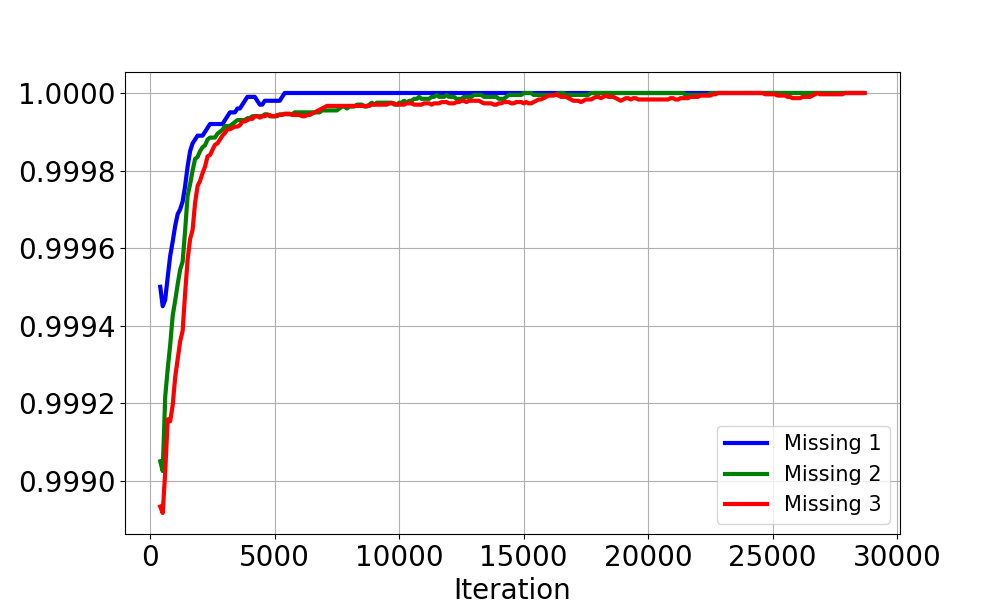}
\caption{Cosine similarity.}\label{fig:ft_cos}
\end{subfigure}
\caption{L2 distance (a) and cosine similarity (b) between generated features and missing features of 1 modality unavailable (T1c, in blue), 2 modalities unavailable (T1c and T2) and 3 modalities unavailable (T1, T1c and T2) over training iterations on BraTS2018.
}
\label{fig:distance_features}
\end{figure}

\noindent\textbf{Examining the Generation of Missing Features.} A potential question regarding MetaKD is its ability to generate suitable features for missing modalities using Eq.~\eqref{eq:ftgen}. To evaluate this, we present the L2 distance (Fig.~\ref{fig:ft_dist}) and cosine similarity (Fig.~\ref{fig:ft_cos}) between the generated feature from Eq.~\eqref{eq:ftgen} and the actual feature from the missing 1 to 3 modalities during training. Notably, distance decreases and similarity increases steadily, showing MetaKD's effectiveness.

\section{Conclusion}

This paper addresses performance drops in multi-modal learning when key modalities are missing. We proposed MetaKD, a novel meta-learning approach that optimizes modality importance dynamically and performs modality-weighted knowledge distillation, distilling knowledge from high- to low-informative modalities. MetaKD achieves strong performance even with missing influential modalities, showing improvements over recent models; for instance, in BraTS2018, we observe Dice score gains of 3.51\% for enhancing tumors, 2.19\% for tumor cores, and 1.14\% for whole tumors. Extensive analysis confirms the effectiveness of learned IWV weights in identifying important modalities.

One potential limitation of our method is the simplistic missing modality feature generation in Eq.~\eqref{eq:ftgen}, which uses the mean feature from available image modalities. 
Preliminary experiments with more advanced feature generation techniques did not yield significant improvements over this simple approach. As a result, we opted against introducing a more complex feature generation method that would add unnecessary complexity without substantially enhancing the final accuracy. Nevertheless, despite its simplicity, Figs.~\ref{fig:ft_dist} and~\ref{fig:ft_cos} demonstrate its effectiveness in generating useful features for MetaKD training.  We still plan, however, to further investigate this point and explore more advanced feature generation methods, such as conditional generative models~\cite{sohn2015learning,mirza2014conditional}.

{
    \small
    \bibliographystyle{ieeenat_fullname}
    \bibliography{main}
}

\input{X_suppl}

\end{document}

%% file: X_suppl.tex
\clearpage
\setcounter{page}{1}
\maketitlesupplementary

\section{Algorithm}

The MetaKD training is presented in Algorithm~\ref{alg}.

\begin{algorithm}[h] \small
	\DontPrintSemicolon
	\SetAlgoLined	
	\KwInput{Multi-modal with missing modality dataset for meta-training and meta-validation $\mathcal{D}=\{\mathcal{D}_t,\mathcal{D}_v\}$, encoder and decoder, \# iterations=$N$}
	\KwOutput{Encoder and decoder parameters $\theta^{*}$ and $\zeta^*$}
	Initialize model parameters $\theta$ and $\zeta$ and meta-parameter $\mathbf{w}$ \\
 
	\While{not converged}{

        \textbf{Meta-validation \& Meta-updating}: \\
        \scalebox{0.8}{$\mathbf{w}^*=\arg\min _{\mathbf{w}} \sum_{(\mathcal{M}, \mathbf{y}) \in \mathcal{D}_v} \ell_{meta} \left(\mathbf{y}, f_{\zeta}\left(\oplus_{i = 1}^{N} \mathbf{w}_i \times \mathbf{f}_{i} \right)\right)$} \\
        Normalize $\mathbf{w}^*$ with Eq. (6) \\
        \textbf{Meta-training}: \\
	    \For{n=1 to N}{
    	Optimize task and MetaKD objectives: \\
       \scalebox{0.85}{$\theta^*,\zeta^* = \arg\min _{\theta,\zeta} \sum_{(\mathcal{M}, \mathbf{y}) \in \mathcal{D}_t} \Big [ \ell_{task} \left(\mathbf{y}, f_{\zeta}\left(\oplus_{i=1}^N 
       \mathbf{f}_{i} \right)\right)$} \\
       \scalebox{0.85}{$\;\;\;\;\;\;\;\;\;\;\;\;\;\;\;\;\;\;\;\;\;\;\;\;\;\;\;\;\;\;\;\;\;\;\;\; + \alpha  \sum_{i,j = 1}^{N}   \frac{\mathbf{w}_i^*}{\mathbf{w}_j^*} \times \ell_{kd}^{(i,j)}\left(  
            \mathbf{f}_{i}, \mathbf{f}_{j} \right)\Big ]
            $}
	    }
        $\theta \leftarrow \theta^*$, $\zeta \leftarrow \zeta^*$ \\
	}	
 \caption{Meta-learned Modality-weighted Knowledge Distillation Framework.}
 \label{alg}
\end{algorithm}

\section{Datasets}

We evaluate our MetaKD on five multi-modal learning with missing modality datasets: BraTS2018, BraTS2019, BraTS2020 for medical image segmentation, ADNI for medical data classification, and Audiovision-MNIST for computer vision classification. 

The BraTS2018, BraTS2019 and BraTS2020 Segmentation dataset~\cite{menze2014multimodal,bakas2018identifying} serves as our benchmarks for multi-modal learning with missing modality in brain tumor sub-region segmentation. These datasets consist of 3D multi-modal brain MRIs, including Flair, T1, T1 contrast-enhanced (T1c), and T2 images, each annotated by experienced imaging experts to provide ground-truth segmentation of specific sub-regions, namely, enhancing tumor (ET), tumor core (TC), and whole tumor (WT).
The BraTS2018 dataset has 285 training/validation cases (210 high-grade and 75 low-grade gliomas) and 66 evaluation cases. Similarly, the BraTS2019 dataset The BraTS2019 dataset consists 335 training/validation cases (259 high-grade gliomas and 76 low-grade gliomas) and 125 evaluation cases; the BraTS2020 dataset includes 369 training/validation cases and 125 evaluation cases. 
Ground-truth annotations are publicly available for the training set, while validation set annotations are hidden, requiring online evaluation\footnote{Online evaluation available at\url{https://ipp.cbica.upenn.edu/}.} for unbiased assessment.

The Alzheimer's Disease Neuroimaging Initiative (ADNI) dataset is a multi-modal resource for studying Alzheimer's disease (AD) progression, including clinical, neuroimaging, genetic, and biospecimen data from 2,939 participants across various AD stages. 
Spanning four phases (ADNI1, ADNIGO, ADNI2, ADNI3), it includes MRI, PET, fMRI, clinical assessments, and genomic data.  However, many participants lack certain modalities, with missing data rates ranging from 30\% to 70\%, making ADNI ideal for missing modality experiments in AD tasks. The dataset is split 70\% for training, 15\% for validation, and 15\% for testing to ensure consistency and reproducibility. 
Certain clinical variables ('PTCOGBEG,' 'PTADDX,' 'PTADBEG') were excluded. Continuous variables were normalized using a MinMax scaler (-1 to 1), while categorical data underwent one-hot encoding. For genetic data, linkage disequilibrium pruning (parameters 50, 5, 0.1) removed redundant SNPs, missing values were imputed with the mean, and MinMax scaling was applied. Biospecimen data followed the same steps as clinical data. For fMRI data, magnetic field inhomogeneity was corrected, and the MUSE framework segmented gray matter. fmriprep was used to standardize fMRI data \cite{esteban2019fmriprep}.

We further conducted missing modality experiments on a general CV dataset --- Audiovision-MNIST~\cite{vielzeuf2018centralnet}, which combines 1,500 audio and image samples. The image data, representing digits 0 to 9 with a resolution of 28$\times$28, comes from the MNIST dataset~\cite{lecun1998gradient}, while the audio data is sourced from the Free Spoken Digits Dataset\footnote{For more info, see \url{https://github.com/Jakobovski/free-spoken-digit-dataset}.}. Following SMIL settings\footnote{Code available at \url{https://github.com/mengmenm/SMIL}.}, we use mel-frequency cepstral coefficients (MFCCs) to transform audio samples into 20$\times$20$\times$1. The dataset is split 60\% for training, 10\% for validation, and 30\% for testing, ensuring consistent evaluation.

\section{Implementation Details and Evaluation Measures}

\noindent\textbf{BraTS:} For all three BraTS dataset experiments, we utilized the 3D UNet architecture as the backbone network, which incorporates 3D convolutions and normalization. Following a standard setup~\cite{havaei2016hemis,dorent2019hetero,chen2019robust,wang2023multi}, the knowledge distillation process of cross-modal features occurs at the bottom of the UNet structure.

To optimize the network, we employed a stochastic gradient descent optimizer with Nesterov momentum~\cite{botev2017nesterov} set at 0.99. The learning rate was initialized to $10^{-2}$ and gradually decreased using the cosine annealing strategy~\cite{loshchilov2016sgdr}. For the optimization of MetaKD weights, we adopt Adam optimizer with $10^{-2}$ as learning rate and $5 \times 10^{-5}$ as the weight decay. Network parameters and MetaKD weights are randomly initialized. We follow the non-dedicated training setting of existing methods~\cite{chen2019robust,zhang2022mmformer,wang2023multi}, where we randomly dropped 0 to 3 modalities to simulate missing-modality scenarios.

The model underwent 115,000 training iterations using the entire training data without model selection. The meta-validation process is conducted every 100 iterations while training. As for our MetaKD loss in Eq. (5), we chose $p=1$ to form an L1 loss. 
For optimization in Eq. (4), we set the hyper-parameter $\alpha=0.1$. 
After training, we performed the official online evaluation using the segmentation masks generated by MetaKD.

\noindent\textbf{ADNI:} The ADNI dataset presents significant class imbalance challenges typical of real-world medical datasets, where control subjects often outnumber patients with cognitive impairment and Alzheimer's disease. This imbalance poses substantial challenges for traditional machine learning approaches, which tend to exhibit bias toward majority classes, potentially leading to poor diagnostic sensitivity for the clinically critical minority classes (MCI and AD patients).

To address this fundamental challenge, we implemented SMOTE (Synthetic Minority Over-sampling Technique) with k-nearest neighbor interpolation (k=5), generating synthetic samples for minority classes while preserving the original imaging data structure. This approach creates realistic synthetic instances that maintain the underlying data distribution characteristics without simple duplication, which could lead to overfitting.

The effectiveness of our class imbalance mitigation is evidenced by the substantial improvements in balanced accuracy metrics compared to standard accuracy measures. For instance, in the Clinical + Genomic combination, while standard accuracy improves by 13.36\%, the balanced accuracy improvement reaches 16.15\%, indicating that MetaKD particularly benefits the underrepresented MCI and AD classes. This pattern is consistent across multiple modality combinations, suggesting that the meta-learning framework, combined with appropriate class balancing, enables more equitable learning across diagnostic categories.

The F1-score improvements are particularly pronounced in scenarios involving clinical and biospecimen data, where class imbalance effects are typically most severe. The Clinical + Biospecimen combination shows a remarkable 29.29\% F1-score improvement, indicating enhanced precision and recall for minority classes that are often missed by conventional approaches.

\noindent\textbf{Audiovision-MNIST:} In the case of model training on the Audiovision-MNIST dataset, we followed the methodology outlined in the SMIL paper~\cite{ma2021smil}. 
More specifically, the image and sound encoders used by SMIL consist of networks with a sequence of convolutional layers and fully connected (FC) layers along with batch normalization and dropout. 
For the remaining model architecture, after fusing the two modality features, 2 FC layers with dropout were adopted for classification. 
The features for the MetaKD loss in Eq.(5) are extracted from the layer just before the FC layers.
 
To optimize the model, we used the Adam optimizer with a weight decay of $10^{-2}$ and an initial learning rate of $10^{-3}$, which was decayed by 10\% every 20 epochs. 
To optimize the importance weight vector, we utilize the Adam optimizer with a learning rate set to $10^{-2}$ and weight decay set to $10^{-2}$. To keep a fair comparison, we train all models (including the compared models and the proposed model) for 60 epochs and drop a certain percentage of the sound modality.

To evaluate the model performance, we employed the Dice score for BraTS2018 and classification accuracy for ADNI and Audiovision-MNIST. All training and evaluation procedures were conducted on a single 3090Ti NVIDIA Graphics Card.

\subsection{Model Performance on BraTS2019 Segmentation}

For the experiments on BraTS2019 dataset, the proposed MetaKD models are compared with WoCM/WoCM+~\cite{}, U-HeMIS (HMIS)~\cite{havaei2016hemis}, U-HVED (HVED)~\cite{dorent2019hetero}, MGP-VAE (MVAE)~\cite{}.

\begin{table*}[h!]
\setlength\tabcolsep{1pt}
\begin{center}
\caption{Model performance comparison on the test set of \textbf{segmentation} Dice score (in \%) on BraTS2019 of \textbf{non-dedicated training}. Our models are compared with WoCM/WoCM+~\cite{zhou2021latent}, U-HeMIS (HMIS)~\cite{havaei2016hemis}, U-HVED (HVED)~\cite{dorent2019hetero}, MGP-VAE (MVAE)~\cite{hamghalam2021modality}. ``Imprv'' denotes the improvement (in percentage) between our proposed MetaKD model and the best of all other models. The best and second best results for each column within a certain type of tumor are in \textbf{bold} and \textit{Italic}, respectively.}\label{tab:brats2019}
\resizebox{1.0\linewidth}{!}{
\begin{tabular}{|cccc|ccccccc|ccccccc|ccccccc|}
\hline
\multicolumn{4}{|c|}{Modalities}              & \multicolumn{7}{c|}{Enhancing Tumor}                                                                                 & \multicolumn{7}{c|}{Tumor Core}                                                                                     & \multicolumn{7}{c|}{Whole Tumor}                                                                                             \\ \hline
Fl        & T1        & T1c       & T2        & WoCM   & WoCM+          & HMIS   & HVED   & MVAE           & MetaKD           & Imprv             & WoCM   & WoCM+          & HMIS   & HVED   & MVAE           & MetaKD           & Imprv            & WoCM           & WoCM+          & HMIS   & HVED   & MVAE           & MetaKD            & Imprv            \\ \hline
$\bullet$ & $\circ$   & $\circ$   & $\circ$   & 13.80  & 17.00          & 24.20  & 24.10  & \textit{25.20} & \textbf{41.46} & \greent{64.54\%}  & 48.10  & 52.00          & 48.80  & 49.90  & \textit{52.70} & \textbf{71.65} & \greent{35.96\%} & 71.90          & 74.30          & 78.90  & 81.00  & \textit{83.30} & \textbf{88.939} & \greent{6.77\%}  \\
$\circ$   & $\bullet$ & $\circ$   & $\circ$   & 0.90   & 1.00           & 11.30  & 13.20  & \textit{14.20} & \textbf{40.76} & \greent{187.01\%} & 0.80   & 2.20           & 37.00  & 36.20  & \textit{39.90} & \textbf{67.98} & \greent{70.36\%} & 8.00           & 10.00          & 53.30  & 51.50  & \textit{53.20} & \textbf{75.349} & \greent{41.63\%} \\
$\circ$   & $\circ$   & $\bullet$ & $\circ$   & 36.20  & 46.80          & 59.50  & 65.10  & \textit{66.40} & \textbf{73.19} & \greent{10.23\%}  & 39.80  & 41.70          & 57.70  & 65.80  & \textit{68.10} & \textbf{79.21} & \greent{16.31\%} & 20.00          & 20.00          & 57.60  & 61.70  & \textit{63.20} & \textbf{76.234} & \greent{20.62\%} \\
$\circ$   & $\circ$   & $\circ$   & $\bullet$ & 3.80   & 3.80           & 22.10  & 29.90  & \textit{30.80} & \textbf{44.30} & \greent{43.84\%}  & 19.60  & 32.60          & 49.30  & 52.80  & \textit{56.20} & \textbf{71.12} & \greent{26.55\%} & 62.90          & 70.30          & 78.00  & 79.90  & \textit{81.10} & \textbf{84.226} & \greent{3.85\%}  \\
$\bullet$ & $\bullet$ & $\circ$   & $\circ$   & 1.80   & 3.00           & 28.20  & 24.10  & \textit{25.10} & \textbf{45.23} & \greent{80.19\%}  & 52.70  & 55.60          & 56.00  & 54.40  & \textit{55.30} & \textbf{74.97} & \greent{35.58\%} & 80.60          & 81.60          & 82.90  & 83.20  & \textit{84.70} & \textbf{89.667} & \greent{5.86\%}  \\
$\bullet$ & $\circ$   & $\bullet$ & $\circ$   & 48.70  & 63.50          & 67.00  & 70.40  & \textit{70.20} & \textbf{72.85} & \greent{3.78\%}   & 72.60  & 74.70          & 66.70  & 72.60  & \textit{74.60} & \textbf{82.32} & \greent{10.35\%} & 81.70          & 81.70          & 82.20  & 84.80  & \textit{85.80} & \textbf{89.946} & \greent{4.83\%}  \\
$\bullet$ & $\circ$   & $\circ$   & $\bullet$ & 6.90   & 11.30          & 26.90  & 33.70  & \textit{35.00} & \textbf{45.45} & \greent{29.87\%}  & 41.10  & 41.80          & 58.00  & 58.90  & \textit{61.00} & \textbf{75.03} & \greent{23.00\%} & 80.60          & 83.90          & 84.80  & 86.50  & \textit{87.90} & \textbf{89.829} & \greent{2.19\%}  \\
$\circ$   & $\bullet$ & $\bullet$ & $\circ$   & 39.70  & 42.30          & 64.30  & 66.60  & \textit{68.00} & \textbf{71.62} & \greent{5.32\%}   & 36.70  & 39.30          & 63.40  & 69.10  & \textit{71.80} & \textbf{80.13} & \greent{11.60\%} & 18.40          & 20.70          & 62.60  & 66.00  & \textit{67.60} & \textbf{79.038} & \greent{16.92\%} \\
$\circ$   & $\bullet$ & $\circ$   & $\bullet$ & 1.00   & 1.20           & 27.10  & 30.30  & \textit{31.80} & \textbf{46.56} & \greent{46.43\%}  & 38.30  & 42.10          & 52.50  & 56.60  & \textit{58.40} & \textbf{74.60} & \greent{27.74\%} & 70.00          & 71.60          & 79.80  & 81.10  & \textit{82.00} & \textbf{86.453} & \greent{5.43\%}  \\
$\circ$   & $\circ$   & $\bullet$ & $\bullet$ & 48.40  & 60.40          & 67.50  & 69.20  & \textit{70.50} & \textbf{72.82} & \greent{3.29\%}   & 63.70  & 66.80          & 68.30  & 73.10  & \textit{75.70} & \textbf{82.63} & \greent{9.15\%}  & 72.30          & 73.50          & 80.10  & 81.80  & \textit{83.10} & \textbf{86.487} & \greent{4.08\%}  \\
$\bullet$ & $\bullet$ & $\bullet$ & $\circ$   & 55.50  & 67.10          & 68.80  & 70.20  & \textit{71.50} & \textbf{73.00} & \greent{2.10\%}   & 75.20  & 75.70          & 69.90  & 73.60  & \textit{75.80} & \textbf{83.40} & \greent{10.03\%} & 83.00          & 83.90          & 83.90  & 85.60  & \textit{87.40} & \textbf{90.114} & \greent{3.11\%}  \\
$\bullet$ & $\bullet$ & $\circ$   & $\bullet$ & 1.70   & 2.90           & 32.30  & 33.30  & \textit{34.50} & \textbf{46.90} & \greent{35.93\%}  & 50.30  & 52.10          & 60.10  & 61.00  & \textit{63.00} & \textbf{76.35} & \greent{21.19\%} & 87.70          & 88.20          & 85.90  & 87.10  & \textit{88.30} & \textbf{90.122} & \greent{2.06\%}  \\
$\bullet$ & $\circ$   & $\bullet$ & $\bullet$ & 52.20  & 64.40          & 68.90  & 70.30  & \textit{71.40} & \textbf{72.31} & \greent{1.28\%}   & 73.60  & 76.60          & 71.60  & 75.20  & \textit{77.40} & \textbf{83.42} & \greent{7.77\%}  & 88.70          & 88.90          & 85.90  & 87.50  & \textit{89.00} & \textbf{90.44}  & \greent{1.62\%}  \\
$\circ$   & $\bullet$ & $\bullet$ & $\bullet$ & 53.10  & 64.00          & 68.70  & 70.40  & \textit{71.60} & \textbf{71.70} & \greent{0.14\%}   & 64.60  & 66.60          & 69.90  & 74.50  & \textit{76.40} & \textbf{83.32} & \greent{9.05\%}  & 73.50          & 74.20          & 81.30  & 82.30  & \textit{83.60} & \textbf{87.244} & \greent{4.36\%}  \\
$\bullet$ & $\bullet$ & $\bullet$ & $\bullet$ & 69.10  & \textit{70.60} & N/A    & N/A    & N/A            & \textbf{73.02} & \greent{3.43\%}   & 76.80  & \textit{77.50} & N/A    & N/A    & N/A            & \textbf{84.09} & \greent{8.51\%}  & \textit{89.70} & \textit{89.70} & N/A    & N/A    & N/A            & \textbf{90.541} & \greent{0.94\%}  \\ \hline
\multicolumn{4}{|c|}{Average}                 & 28.85  & 34.62          & 45.49  & 47.91  & \textit{49.01} & \textbf{59.41} & \greent{21.22\%}  & 50.26  & 53.15          & 59.23  & 62.41  & \textit{64.74} & \textbf{78.01} & \greent{20.5\%}  & 65.93          & 67.50          & 76.94  & 78.57  & \textit{80.01} & \textbf{86.31}  & \greent{7.87\%}  \\ \hline
\multicolumn{4}{|c|}{p-value}                 & 7.5e-8 & 1.4e-5         & 2.6e-5 & 1.7e-4 & 4.2e-4         & N/A            & N/A               & 1.2e-5 & 2.3e-5         & 2.6e-9 & 5.0e-7 & 2.1e-6         & N/A            & N/A              & 1.5e-3         & 2.5e-3         & 2.4e-5 & 2.0e-4 & 9.4e-4         & N/A             & N/A              \\ \hline
\end{tabular}
}\end{center}
\end{table*}

In the experiments on BraTS2019 dataset, the proposed MetaKD model demonstrates significant improvements over existing methods across various modalities and tumor types. For instance, in the segmentation of the enhancing tumor, MetaKD achieves an average Dice score of 59.41, surpassing the best competing model by 21.22\%. Similarly, for tumor core segmentation, MetaKD outperforms previous models by achieving an average Dice score of 78.01, representing a 20.5\% improvement over the best alternative. The comparison highlights MetaKD’s superiority, respectively. In the segmentation of the whole tumor, MetaKD also shows a clear advantage, with an average Dice score of 86.31. This score is notably higher than that of the next best model, which is only 80.01, reflecting a 7.87\% improvement. These findings confirm MetaKD’s overall superior performance and its potential as a robust tool for medical image segmentation, outperforming several established models in various aspects of tumor segmentation.

\subsection{Model Performance on BraTS2020 Segmentation}

On the dataset of BraTS2020, the compared methods include U-HeMIS (abbreviated as HMIS in the table)~\cite{havaei2016hemis}, U-HVED (HVED)~\cite{dorent2019hetero}, Robust-MSeg (RSeg)~\cite{chen2019robust}, SFusion~\cite{liu2023sfusion} and RFNet~\cite{ding2021rfnet}.

\begin{table*}[h!]
\setlength\tabcolsep{1pt}
\begin{center}
\caption{Model performance comparison on the test set of \textbf{segmentation} Dice score (in \%) on BraTS2020 of \textbf{non-dedicated training}. Our models are compared with U-HeMIS (HMIS)~\cite{havaei2016hemis}, U-HVED (HVED)~\cite{dorent2019hetero}, Robust-MSeg (RSeg)~\cite{chen2019robust}, SFusion~\cite{liu2023sfusion} and RFNet~\cite{ding2021rfnet}. ``Imprv'' denotes the improvement (in percentage) between our proposed MetaKD model and the best of all other models. The best and second best results for each column within a certain type of tumor are in \textbf{bold} and \textit{Italic}, respectively.}\label{tab:brats2020}
\resizebox{1.0\linewidth}{!}{
\begin{tabular}{|cccc|ccccccc|ccccccc|ccccccc|}
\hline
\multicolumn{4}{|c|}{Modalities}              & \multicolumn{7}{c|}{Enhancing Tumor}                                                                                 & \multicolumn{7}{c|}{Tumor Core}                                                                                     & \multicolumn{7}{c|}{Whole Tumor}                                                                                             \\ \hline
Fl        & T1        & T1c       & T2        & HMIS   & HVED   & RbSeg  & SFusion & RFNet          & MetaKD           & Imprv          & HMIS   & HVED   & RbSeg  & SFusion & RFNet          & MetaKD           & Imprv         & HMIS   & HVED   & RbSeg  & SFusion & RFNet          & MetaKD           & Imprv         \\ \hline
$\bullet$ & $\circ$   & $\circ$   & $\circ$   & 9.00   & 20.87  & 34.68  & 34.40   & \textit{38.15} & \textbf{42.41} & \greent{11.16\%} & 24.97  & 51.15  & 60.72  & 52.84   & \textit{69.19} & \textbf{72.15} & \greent{4.27\%} & 52.29  & 82.69  & 82.87  & 83.97   & \textit{87.32} & \textbf{88.96} & \greent{1.88\%} \\
$\circ$   & $\bullet$ & $\circ$   & $\circ$   & 16.53  & 12.33  & 28.99  & 29.71   & \textit{37.30} & \textbf{42.72} & \greent{14.53\%} & 42.42  & 36.73  & 54.30  & 53.86   & \textit{66.02} & \textbf{68.27} & \greent{3.40\%}  & 63.01  & 54.93  & 71.41  & 69.11   & \textbf{77.16} & \textit{76.83} & \redt{-0.43\%}  \\
$\circ$   & $\circ$   & $\bullet$ & $\circ$   & 63.24  & 66.59  & 67.91  & 71.94   & \textit{74.85} & \textbf{77.64} & \greent{3.73\%}  & 69.41  & 73.01  & 76.68  & 75.63   & \textit{81.51} & \textbf{82.03} & \greent{0.64\%} & 64.58  & 68.54  & 71.39  & 69.75   & \textit{76.77} & \textbf{77.03} & \greent{0.34\%} \\
$\circ$   & $\circ$   & $\circ$   & $\bullet$ & 31.43  & 28.70  & 36.46  & 35.87   & \textit{46.29} & \textbf{46.65} & \greent{0.78\%}  & 54.22  & 57.43  & 61.88  & 61.99   & \textit{71.02} & \textbf{74.86} & \greent{5.40\%}  & 79.85  & 80.75  & 82.20  & 79.61   & \textit{86.05} & \textbf{89.84} & \greent{4.40\%} \\
$\bullet$ & $\bullet$ & $\circ$   & $\circ$   & 13.99  & 22.53  & 39.67  & 38.22   & \textit{40.98} & \textbf{43.57} & \greent{6.32\%}  & 41.58  & 55.10  & 68.18  & 62.31   & \textit{73.07} & \textbf{74.86} & \greent{2.45\%} & 65.29  & 85.01  & 88.10  & 86.39   & \textit{89.73} & \textbf{89.84} & \greent{0.12\%} \\
$\bullet$ & $\circ$   & $\bullet$ & $\circ$   & 68.31  & 69.53  & 70.78  & 40.38   & \textit{76.67} & \textbf{77.77} & \greent{1.44\%}  & 70.86  & 76.86  & 81.85  & 66.67   & \textit{84.65} & \textbf{85.13} & \greent{0.57\%} & 69.37  & 86.13  & 87.33  & 81.78   & \textit{89.89} & \textbf{90.27} & \greent{0.42\%} \\
$\bullet$ & $\circ$   & $\circ$   & $\bullet$ & 28.91  & 30.48  & 42.19  & 41.46   & \textbf{49.32} & \textit{49.15} & \redt{-0.35\%}   & 55.89  & 61.87  & 68.20  & 66.38   & \textit{74.14} & \textbf{75.00} & \greent{1.16\%} & 81.56  & 87.40  & 88.09  & 87.50   & \textit{89.87} & \textbf{90.10} & \greent{0.25\%} \\
$\circ$   & $\bullet$ & $\bullet$ & $\circ$   & 70.71  & 67.82  & 70.11  & 74.90   & \textbf{78.01} & \textit{77.51} & \redt{-0.64\%}   & 75.59  & 76.49  & 80.28  & 80.35   & \textit{83.4}  & \textbf{83.54} & \greent{0.17\%} & 72.50  & 71.58  & 76.84  & 75.30   & \textit{81.12} & \textbf{86.10} & \greent{6.14\%} \\
$\circ$   & $\bullet$ & $\circ$   & $\bullet$ & 28.58  & 28.73  & 39.92  & 40.38   & \textit{45.65} & \textbf{46.03} & \greent{0.82\%}  & 56.38  & 59.29  & 66.46  & 66.67   & \textit{73.13} & \textbf{75.04} & \greent{2.60\%}  & 82.31  & 81.58  & 85.53  & 81.78   & \textit{87.73} & \textbf{90.03} & \greent{2.62\%} \\
$\circ$   & $\circ$   & $\bullet$ & $\bullet$ & 70.30  & 68.74  & 71.42  & 74.74   & \textit{75.93} & \textbf{77.80} & \greent{2.46\%}  & 77.60  & 77.85  & 82.44  & 81.48   & \textit{83.45} & \textbf{83.86} & \greent{0.49\%} & 84.45  & 83.37  & 85.97  & 84.27   & \textit{87.74} & \textbf{90.24} & \greent{2.85\%} \\
$\bullet$ & $\bullet$ & $\bullet$ & $\circ$   & 70.80  & 71.32  & 71.77  & 75.44   & \textit{76.81} & \textbf{77.48} & \greent{0.87\%}  & 75.07  & 79.51  & 82.76  & 82.04   & \textit{85.07} & \textbf{85.16} & \greent{0.10\%} & 73.31  & 87.10  & 88.87  & 88.04   & \textbf{90.69} & \textit{90.43} & \redt{-0.29\%}  \\
$\bullet$ & $\bullet$ & $\circ$   & $\bullet$ & 29.53  & 30.60  & 43.90  & 43.53   & \textit{49.92} & \textbf{50.14} & \greent{0.44\%}  & 57.40  & 63.46  & 70.46  & 68.76   & \textit{75.19} & \textbf{75.5}  & \greent{0.43\%} & 83.03  & 88.07  & 89.24  & 87.63   & \textbf{90.60} & \textit{90.26} & \redt{-0.38\%}  \\
$\bullet$ & $\circ$   & $\bullet$ & $\bullet$ & 71.36  & 69.84  & 71.17  & 74.91   & \textit{77.12} & \textbf{77.87} & \greent{0.97\%}  & 77.69  & 78.68  & 81.89  & 82.06   & \textit{84.97} & \textbf{85.22} & \greent{0.30\%}  & 84.64  & 88.33  & 88.68  & 89.11   & \textit{90.68} & \textbf{90.72} & \greent{0.04\%} \\
$\circ$   & $\bullet$ & $\bullet$ & $\bullet$ & 71.67  & 69.74  & 71.87  & 74.78   & \textit{76.99} & \textbf{77.31} & \greent{0.42\%}  & 79.05  & 79.99  & 82.85  & 82.32   & \textit{83.47} & \textbf{84.42} & \greent{1.14\%} & 85.19  & 84.27  & 86.63  & 84.59   & \textit{88.25} & \textbf{89.23} & \greent{1.11\%} \\
$\bullet$ & $\bullet$ & $\bullet$ & $\bullet$ & 71.49  & 70.50  & 71.52  & 73.76   & \textit{78.00} & \textbf{78.01} & \greent{0.01\%}  & 78.58  & 80.40  & 82.87  & 82.18   & \textit{85.21} & \textbf{85.79} & \greent{0.68\%} & 85.19  & 88.81  & 89.47  & 88.93   & \textit{91.11} & \textbf{91.33} & \greent{0.24\%} \\ \hline
\multicolumn{4}{|c|}{Average}                 & 47.73  & 48.55  & 55.49  & 57.32   & \textit{61.47} & \textbf{62.80} & \greent{2.17\%}  & 65.45  & 67.19  & 73.45  & 71.86   & \textit{78.23} & \textbf{79.39} & \greent{1.48\%} & 75.10  & 81.24  & 84.17  & 82.89   & \textit{86.98} & \textbf{88.08} & \greent{1.26\%} \\ \hline
\multicolumn{4}{|c|}{p-value}                 & 9.5e-6 & 1.1e-6 & 4.7e-9 & 1.9e-3  & 4.7e-3         & N/A            & N/A            & 2.8e-5 & 1.2e-5 & 3.3e-5 & 3.8e-5  & 7.5e-4         & N/A            & N/A           & 2.8e-5 & 1.1e-4 & 1.3e-5 & 3.0e-6  & 1.0e-2         & N/A            & N/A           \\ \hline
\end{tabular}
}\end{center}
\end{table*}

The table~\ref{tab:brats2020} presents a comprehensive comparison of the proposed MetaKD method with several existing approaches across different evaluation metrics. Similarly to the performance of the model on the BraTS 2018 and BraTS2019 data, the proposed MetaKD model receives significantly better results than its counterparts.

The proposed MetaKD consistently achieves superior performance, yielding the highest dice score across almost all tested scenarios. Notably, the improvement in performance, denoted as "Imprv," is significant for most missing modality combinations, with MetaKD outperforming the next best method (RFNet) in the majority of cases (43 out of 48, including the average score). The improvements in some cases are over 5\%, especially on those single modality available and without the strongest modality scenarios. This performance highlights the robustness and effectiveness of MetaKD. The average performance metrics further emphasize MetaKD's advantage (2.17\% improvement on ET, 1.48\% improvement on TC and 1.26\% improvement on WT), showing substantial gains over the other methods, with statistically significant improvements as indicated by p-values well below conventional thresholds (0.05). These results strongly suggest that MetaKD offers a more accurate and reliable solution for the tasks at hand, setting a new benchmark in the field.

\section{Normalization Functions for Eq. (6)}

We tried different IWV normalization functions in Eq. (6), with results in Table~\ref{tab:norm}.
Note that the softmax normalization leads to the best results.

\begin{table}[htbp]
\caption{Segmentation Dice scores (in \%) to experiment with different normalization functions, including ReLU~\cite{nair2010rectified}, Sigmoid function and Softmax function, adopted for IWV learning in Eq. (6), using Flair, T1 and T1c for testing on BraTS2018.
}
\begin{center}
\resizebox{0.9\linewidth}{!}{
\begin{tabular}{|c|ccc|}
\hline
Normalization   & Enhancing Tumor & Tumor Core & Whole Tumor \\ \hline
$\text{ReLU}(\cdot)$ & 71.20                       & 81.99                       & 90.31                       \\
$\text{Sigmoid}(\cdot)$  & 73.99                       & 80.18                       & 87.74                       \\
$\text{Softmax}(\cdot)$    & 77.66                       & 85.36                       & 90.70                       \\ \hline
\end{tabular}
}\end{center}
\label{tab:norm}
\end{table}

\section{More Visualizations}

\begin{figure}[htbp]
\centering
\includegraphics[width=0.45\textwidth]{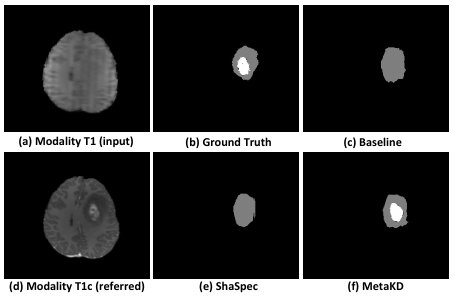}
\caption{Additional visualization of the segmentation results produced by different models using T1 only for testing on BraTS2018.
}
\label{fig:vis-seg}
\end{figure}

\begin{figure}[htbp]
\centering
\includegraphics[width=0.45\textwidth]{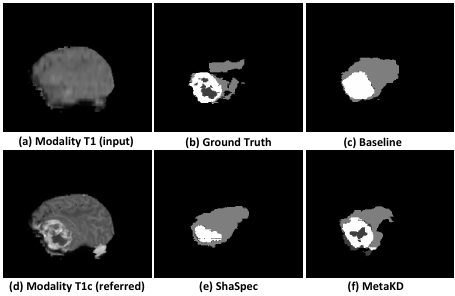}
\caption{Additional visualization of the segmentation results produced by different models using T1 only for testing on BraTS2018.
}
\label{fig:vis-seg2}
\end{figure}

We provide additional segmentation results of different models on BraTS2018 dataset. In Figure~\ref{fig:vis-seg}, while both the baseline model and ShaSpec exhibited shortcomings in detecting the ET tumor (in white), MetaKD demonstrated the capability of successfully identifying it. A similar situation is showed in Figure~\ref{fig:vis-seg2}, where our MetaKD achieves qualitatively better segmentation results on all three tumor regions, compared to the Baseline model or ShaSpec. 

\begin{figure}
    \centering
    \includegraphics[width=0.9\linewidth]{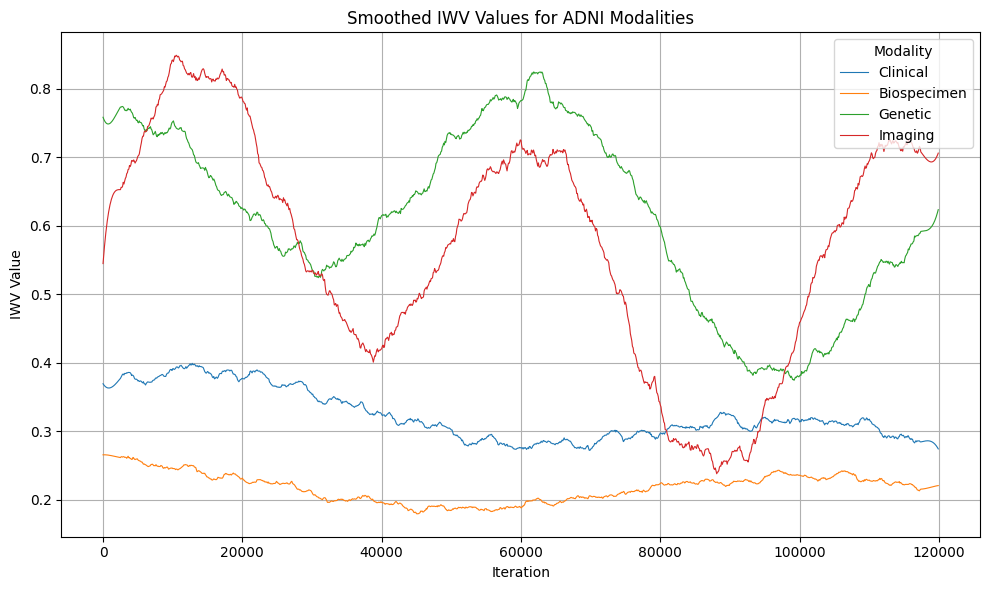}
    \caption{Smoothed Importance Weight Values (IWV) Across Iterations for ADNI Modalities in MetaKD Model with Window Size 50}
    \label{fig:ADNI_IWV}
\end{figure}

Regarding ADNI experimentation, Figure ~\ref{fig:ADNI_IWV} illustrates the smoothed IWV values for four ADNI modalities (Clinical, Biospecimen, Genetic, Imaging) across iterations. The figure shows several key trends that are consistent with the quantitative results. Both imaging and Genetic modalities show dominant values stressing their importance in AD diagnosis. On the other hand, the clinical modality maintains a steady, moderate importance throughout, with minimal fluctuations. Finally, as expected from the results, the biospecimen exhibits the lowest importance among the modalities, remaining relatively flat with a slight downward drift across iterations. These trends suggest that imaging and genetic modalities contribute more significantly and variably to the model, while clinical and biospecimen provide stable but less impactful contributions.

